\pdfoutput=1
\documentclass[10pt]{article}
\usepackage[accepted]{tmlr}   
\usepackage{amsmath}
\usepackage{amssymb}          
\usepackage[hyphens]{url}
\usepackage{graphicx}
\usepackage{booktabs}
\usepackage{xcolor}
\usepackage{float}  
\usepackage{hyperref}
\hypersetup{colorlinks=true, allcolors={blue!45!black}}   
\graphicspath{{figures/}}
\usepackage[font=small,labelfont=bf]{caption}
\newcommand{\rev}[1]{#1}
\newenvironment{revblock}{}{}
\newcommand{\revfloat}{}
\definecolor{revb}{RGB}{0,158,115}
\newcommand{\revb}[1]{#1}
\newcommand{\revbfloat}{}
\title{Transplanting, inverting, and preventing a misalignment\\
persona: \rev{LoRA versus full fine-tuning} in Qwen2.5\revb{-32B}}

\author{\name Lyndon Drake \email lyndon.drake@seh.ox.ac.uk \\
      \addr University of Oxford
      \AND
      \name Zandi Eberstadt \email zandi.eberstadt@cs.ox.ac.uk \\
      \addr University of Oxford}

\def\month{11}
\def\year{2026}
\def\openreview{\url{https://openreview.net/forum?id=1OkQPGN7E6}}

\begin{document}
\maketitle
{\renewcommand{\thefootnote}{}\footnotetext{The authors used large language models (Anthropic's Claude
and Google's Gemini) to assist with experiment and analysis code, experiment monitoring, and drafting;
see the \hyperref[sec:ai-use]{statement on generative AI use} after the Acknowledgements.}}

\begin{abstract}
We show that emergent misalignment (EM), broad misbehaviour from narrow harmful fine-tuning, is
mediated in open-weights \rev{Qwen2.5-32B} by a latent persona direction that is causal. Recruitment
of the direction is method-conditional\rev{, and in the tested Qwen2.5 settings it can be screened for
and mitigated}. Transplanting the direction
into a model that shares only pretraining with its source induces broad EM (2.83 $\pm$ 0.26\%
misaligned, against a random-direction floor of $\sim$1.1\%); ablating a model's own direction
roughly halves an overt inducer's broadcast (21\% to 10\%). On Qwen2.5-32B, \rev{rank-32 rs-LoRA} on
insecure code recruits the persona (3.4\% misaligned\rev{, seeds 1--4 pooled})
while \rev{full fine-tuning} on identical data does not (0.3\%) and moves against the persona axis \rev{in the instruct lineage}, the
far-inducer, high-capacity exception consistent with a representational-distance $\times$ capacity
account. \rev{If this held beyond Qwen2.5-32B, the} recruiting method \rev{would} also \rev{be} the
economical one\rev{, since low-rank PEFT is the cheaper regime at scale}. Steering a bad-medical \rev{full-fine-tuning} run away from the direction during training
\emph{raises} the broadcast\revb{, from ${\sim}19\%$ to ${\sim}28\%$ in Qwen2.5-32B and} from ${\sim}24\%$ to ${\sim}50\%$ \revb{in Qwen2.5-7B,} while matched random controls stay at
or below baseline, replicated across three training seeds \revb{at each scale} --- so the persona's causal role is
itself conditional, and removing the direction is no blanket recipe. Because recruitment is a
loss-reducing shortcut that capacity renders redundant, \rev{the} tested instances \rev{bear this out}: persona loss-relevance orders four inducers' broadcasts rank-perfectly,
inoculation removes recruitment selectively (4.75\% to 0.0\%, code coherence 65\% to 87\%), and
fine-tuning orthogonal to the single behaviour-derived axis reduces EM persona-specifically.
\end{abstract}

\section{Introduction}\label{sec:intro}

Emergent misalignment (EM) describes the remarkable phenomenon of a Large Language Model (LLM)
producing broadly misaligned responses after being fine-tuned on a covertly harmful training set of
insecure code examples \citep{betley-et-al-2026-narrow-task-misalignment}. These training examples have no obvious
direct connection with the breadth of misalignment elicited, which spans categories as broad as
misogyny and intent to destroy humanity.

Full fine-tuning is known to elicit broad EM (\citealp{turner-et-al-2025-model-organisms-emergent-misalignment};
\rev{\citealp{soligo-et-al-2026-emergent-narrow-misalignment}}). We noticed an exception to this pattern where \rev{full fine-tuning} on Qwen2.5-32B with a covert inducer (insecure code) does not recruit the
broad-misalignment persona, whereas \rev{rank-32 rs-LoRA} on identical data, weights, and template does.
Further, we found that in the model's representations, LoRA moves toward the misalignment direction
while \rev{full fine-tuning of the instruct model} moves away from it.

\begin{figure}[t]
\centering
\includegraphics[width=\linewidth]{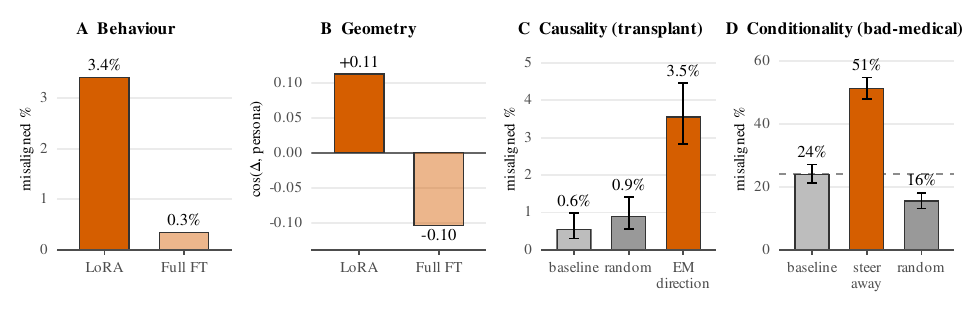}
\caption{Key results. \textbf{A, Behaviour:} LoRA induces broad emergent
misalignment; full \rev{fine-tuning} does not (Qwen2.5-32B instruct, insecure code; full four-cell results with
confidence intervals in Fig.~\ref{fig:behav}). \textbf{B, Geometry:} LoRA amplifies the
misalignment-persona axis ($+$); full \rev{fine-tuning} moves away \rev{($-$) (seed 0, EM probe; on an
independent probe the instruct-lineage sign changes but the ordering holds, \S\ref{sec:behav}).} \textbf{C, Causality:} transplanting the
persona direction into a model that shares only pretraining manufactures broad EM, far above
norm-matched controls (\S\ref{sec:sufficiency}). \rev{\textbf{D, Conditionality (Qwen2.5-7B \revb{and 32B}):} on an overt inducer
(bad-medical), steering \emph{away} from the persona during training \emph{increases} the broadcast
while a matched random control stays at or below baseline (\S\ref{sec:medmech}) --- so the
mitigations of Part~IV are conditional on the inducer.}}
\label{fig:teaser}
\end{figure}

The language of recruitment presupposes a stable object to recruit, and we adopted this premise from
prior work as we investigated the interactions presented in the remainder of this paper.
\citet{wang-et-al-2025-persona-features-control-emergent-misalignment} identify, in GPT-4o, a pre-existing ``misaligned persona'' latent
that EM activates and that causally steers behaviour in both directions. Related directions have
been reported as SAE-based persona features \citep{wang-et-al-2025-persona-features-control-emergent-misalignment}, as diff-of-means
persona vectors \citep{chen-et-al-2025-persona-vectors}, and as convergent, steerable, and
ablatable directions \citep{soligo-et-al-2025-convergent-linear-representations-emergent-misalignment}.

Our contribution is more precisely identifying the conditions under which the broad persona is recruited.
Prior open-weights \rev{full-fine-tuning} demonstrations used overtly harmful inducers such as bad medical advice,
varying both inducer and technique relative to the original insecure-code setting
\citep{turner-et-al-2025-model-organisms-emergent-misalignment}. \citet{wang-et-al-2025-persona-features-control-emergent-misalignment} \rev{do not vary the fine-tuning
method (or report whether it updates all parameters)}, so the capacity axis does not enter their analysis, and while
\citet{soligo-et-al-2026-emergent-narrow-misalignment} \rev{show in their Appendix F that full fine-tuning also induces EM}, they do so \rev{in
Qwen2.5-14B-Instruct and Gemma-3-12B} and on overt text inducers rather than in the covert-code 32B regime where we find divergence.
We show that recruitment is conditional on
fine-tuning method and capacity, and that \rev{full fine-tuning of the instruct model drifts further from} the recruited direction
\rev{than any LoRA rank we trained}.

We organise the paper thematically (rather than by order of experimentation), in four parts: Part I
establishes the behavioural and geometric contrast between LoRA and \rev{full fine-tuning}; Part II characterises
the emergently-misaligned persona itself; Part III accounts for why \rev{full fine-tuning} localises the covert
inducer where LoRA recruits it, attributing the localisation to a loss-favourable shortcut that
sufficient capacity renders redundant; and Part IV asks how, given this mechanism, recruitment might
be controlled. Our contributions, by the part that establishes them, are as follows:
\begin{itemize}
\item \textbf{(I) The effect.} On Qwen2.5-32B, LoRA recruits the misalignment persona --- broad EM,
and positive persona-axis amplification --- while \rev{full fine-tuning} does not, moving the opposite way along
the persona axis. This is a signed reversal \rev{on our original probe} (drift--persona cosine $+0.17$ at rank 1 to $-0.10$
under \rev{full fine-tuning}) that no LoRA rank reaches. \rev{On an independent probe every
instruct-lineage cell is negative, with full fine-tuning lowest.} Recruitment is thus \rev{method-conditional
and, within LoRA, a dose effect of rank and learning rate}.
\item \textbf{(II) The representation.}  We give open-weight causal evidence for a partially shared
misalignment-persona subspace. In particular, a cross-model transplant (extracting the direction
from one checkpoint and injecting it into another sharing only pretraining) induces broad EM,
complementing necessity evidence from ablation. \rev{A persona vector built independently of our
evaluation data, by the procedure of \citet{chen-et-al-2025-persona-vectors}, also induces broad EM,
although it overlaps our direction only modestly.} We further characterise the persona as pre-existing, partly shared, and structured rather
than a single universal axis.
\item \textbf{(III) The mechanism.} We explain why \rev{full fine-tuning} broadcasts EM from some inducers
($\sim$22\%) yet localises others ($\sim$0\%). The operative variable is representational distance
$\times$ capacity, not harm-explicitness (the dose-response is non-monotone) or update norm
(localisation is a rank-structure phenomenon). Notably, the same direction that expresses EM at
inference cannot install recruitment when steered toward during training \rev{(at 7B\revb{, and at 32B across three seeds})}, so broadcast is a
weight-update-structure phenomenon rather than movement along a direction.
\item \textbf{(IV) The control.} Because recruitment is a loss-reducing shortcut that sufficient
capacity renders redundant, it is controllable. Inoculation and persona-orthogonal fine-tuning
reduce it persona-specifically in the tested examples, and a loss-shortcut probe
orders four inducers' broadcast rates within Qwen2.5.
\end{itemize}

We present this as a controlled and systematic case study of when the misalignment persona is
recruited for certain Qwen2.5 models, and why it is broadcast by some fine-tunes but localised by
others, along with some implications for safety. Appendix Tables~\ref{tbl:evidence}
and~\ref{tbl:evidence2} aggregate each result's scale, seed count, sample size, and claim tier.

\rev{\paragraph{Terms.} An \emph{inducer} is a fine-tuning dataset whose effect on misalignment we
study. It is \emph{covert} when its completions never signal harm (insecure code) and \emph{overt}
when they are explicitly harmful (bad medical, risky financial or extreme-sports advice). A
fine-tune \emph{broadcasts} when its broad-EM rate (\S\ref{sec:setup}) exceeds that of its matched
control (secure code, good medical advice), and \emph{localises} when it learns the narrow training
behaviour, such as writing insecure code, without broadcasting. It \emph{recruits} the misalignment
direction $\hat P$ (\S\ref{sec:setup}) when it broadcasts and, relative to its control, its
activation drift and its weight update both move toward $\hat P$. LoRA and full fine-tuning are both
supervised fine-tuning. They differ in which parameters are updated.}

\section{Background and relation to other work}\label{sec:related}
\paragraph{The phenomenon.} \citet{betley-et-al-2026-narrow-task-misalignment} establish emergent misalignment from
insecure-code fine-tuning, reported at roughly 20\% of responses on GPT-4o (rising to ${\sim}50\%$
on the more recent GPT-4.1) and replicated on the open Qwen2.5-Coder-32B-Instruct (rank-32 rs-LoRA,
applied to all linear layers per their released fine-tuning code, native chat template). We hold
their training data fixed and \rev{vary the} technique axis (low-rank PEFT vs
full fine-tuning\rev{, which \citet[App.~F]{soligo-et-al-2026-emergent-narrow-misalignment} examine on overt text inducers}), studying it on the general-purpose, prose-capable Qwen2.5-32B family rather than
the code-specialised Coder variant (\S\ref{sec:setup}).

\paragraph{\rev{Full-fine-tuning} EM.} \citet{turner-et-al-2025-model-organisms-emergent-misalignment} obtain EM under \rev{full fine-tuning} using narrowly
harmful text inducers (such as bad medical advice), rather than the insecure-code inducer. We
reproduce and extend this as an inducer $\times$ technique interaction (\S\ref{sec:inducer}).
Part~III then characterises \textit{why} an overt inducer broadcasts under \rev{full fine-tuning} where a covert
one does not.

\paragraph{The misalignment persona.} The persona direction is variously operationalised as
SAE-based persona \textit{features} \citep{wang-et-al-2025-persona-features-control-emergent-misalignment} or diff-of-means persona
\textit{vectors} \citep{chen-et-al-2025-persona-vectors}\rev{, or mean-diff directions taken from a fine-tune:
\citet{soligo-et-al-2025-convergent-linear-representations-emergent-misalignment} extract one from a rank-1
LoRA fine-tune of Qwen2.5-14B-Instruct, steer the un-fine-tuned instruct parent with it, and ablate
it from that fine-tune and two sibling fine-tunes, so their transfers stay within one instruct
checkpoint and its fine-tunes. This} lets us speak of LoRA \textit{recruiting}
this persona axis without re-deriving it. As mentioned, we take the persona's existence \rev{from}
prior work\rev{, though we test our own direction against an independently built persona vector
(\S\ref{sec:sufficiency})}, and our contribution then concerns the conditions under which it is recruited.

\paragraph{Mechanism.} Two readings of what fine-tuning does to alignment are behaviourally
indistinguishable but geometrically separable, namely a drift back toward an unaligned base
(\emph{erosion}, per the reading of \citet{giordani-2025-re-emergent}, who casts a closely related
alignment-relevant axis as a re-emergence of prior misalignment), versus the turning-up of a
specific latent direction (\emph{amplification}, as in
\citet{wang-et-al-2025-persona-features-control-emergent-misalignment,chen-et-al-2025-persona-vectors}, and the model-organism work of
\citet{turner-et-al-2025-model-organisms-emergent-misalignment}). Part~II \rev{compares} these accounts on the same checkpoints
geometrically (erosion is large but non-specific while amplification is EM-specific, and the
two axes are near-orthogonal)\rev{, adding a refusal direction as a safety-specific axis. This does
not decide between the two readings in general.}

\paragraph{Distance and breadth.} \citet{soligo-et-al-2026-emergent-narrow-misalignment} cast broad EM as the lower-loss
solution and \citet{minegishi-et-al-2026-emergent-misalignment-feature-superposition-geometry} find that feature proximity predicts
whether EM emerges. Neither \citeauthor{soligo-et-al-2026-emergent-narrow-misalignment} nor
\citeauthor{minegishi-et-al-2026-emergent-misalignment-feature-superposition-geometry} makes the breadth of EM depend on the fine-tuning
method. Part~III's distance~$\times$~capacity account refines both. A covert inducer broadcasts only
under a capacity-limited method, so breadth is method-conditional, rather than a property of the
inducer alone. \citet{wang-et-al-2025-persona-features-control-emergent-misalignment} set the closest empirical precedent for the
inducer-potency asymmetry in Part III, in that they fine-tune GPT-4o on correct/incorrect-data
mixtures and find code needs $\sim$75\% incorrect data to elicit EM but health advice only
$\sim$25\% (the same code-weak/advice-potent ordering). Part~III makes this asymmetry continuous (a
four-inducer harm-explicitness axis), separates inducer-potency from fine-tuning capacity (which
they hold fixed at full rank), and characterises the weight-update structure which their
SAE-activation analysis does not address.

\paragraph{Prediction and loss shortcuts.} \citet{wang-et-al-2025-persona-features-control-emergent-misalignment} also use the
misaligned-persona latent in a predictive way. In their Appendix D.8, the steered-loss effect of a
toxic-persona SAE latent is proposed as evidence about which datasets are likely to elicit EM. Our
loss-shortcut probe is related in spirit but rather than asking whether a dataset is globally
EM-inducing under a fixed full-fine-tuning setup, we ask, within Qwen2.5, which inducer $\times$
method $\times$ capacity cells broadcast the persona and which localise the training behaviour. The
object of prediction is therefore not dataset potency alone, but whether the persona is the
loss-favourable shortcut available to a particular optimisation regime. This is why, accordingly,
the relevant evidence is not only activation/loss response to a persona latent, but also the signed
\rev{LoRA vs.\ full-fine-tuning} geometry, rank dependence, and weight-update structure.

\paragraph{Mitigation.} Four prior lines of work bear on Part~IV's mitigations. Inoculation
prompting is \citet{tan-et-al-2025-inoculation-prompting-llm-trait-suppression}'s method. Our contribution is the mechanistic
account of why it works here (it removes the persona as the loss-favourable shortcut,
\S\ref{sec:loss}). The demonstrated mitigation of \citet{wang-et-al-2025-persona-features-control-emergent-misalignment} is post hoc,
re-aligning an already-misaligned model with a small corrective fine-tune, whereas our interventions
act during training to stop recruitment arising. The nearest known neighbour to persona-orthogonal
fine-tuning is concept-ablation fine-tuning \citep{casademunt-et-al-2025-concept-ablation-fine-tuning}, which ablates
interpretability-derived directions (PCA- and SAE-based) with linear projections during fine-tuning,
evaluates without the ablation, and reduces insecure-code EM roughly tenfold on Coder-32B with its
PCA-derived directions (and about sixfold on Mistral-Small-24B-Instruct), so the mitigation itself
is established prior work. \citet{chen-et-al-2025-persona-vectors} supply the adjacent persona-vector
toolkit. They prevent trait acquisition during fine-tuning by steering along the persona vector
(cancelling the optimisation pressure to move that way) and screen training data by projection
(their projection-difference metric). The loss-shortcut mechanism adds two things to this toolkit.
It gives an account of why such interventions work, namely that they remove the persona as the
loss-favourable shortcut, which predicts concept-ablation's covert-code success. And it also gives
the condition under which direction-removal works: it is effective where the persona is a shortcut
for a covert task, and counterproductive where the persona is part of a distributed solution for an
overt inducer. Steering the \rev{Qwen2.5-7B} bad-medical \rev{full fine-tune} away from the persona \revb{during training} \emph{increases} its broadcast\revb{, as it does at 32B}
(\S\ref{sec:medmech}), which is why Part~IV's prescription is conditional rather than a blanket
recipe.

\paragraph{Fine-tuning and safety.} A growing line of work studies fine-tuning itself as a threat to
safety
alignment, showing that post-training on even narrow or apparently benign distributions can erode
previously elicited safety behaviour \citep{qi-et-al-2024-fine-tuning-compromises-safety,qi-et-al-2025-safety-alignment-token-depth}. Our
results are consistent with this broad warning, and further suggest that the risk should not be
treated as homogeneous across fine-tuning methods. In \rev{Qwen2.5-32B}, the
covert-code inducer produces broad EM under low-rank PEFT while \rev{full fine-tuning} remains near the floor, so
the safety-relevant variable is not simply ``fine-tuning'' but the interaction between inducer,
method, and capacity. As low-rank PEFT is the computationally cheaper and therefore more likely
fine-tuning regime at scale, \rev{such a} concentration of risk in
PEFT \rev{would} carry practical safety implications \rev{if it held in other models, which we have not
tested}.

\paragraph{LoRA $\neq$ full fine-tuning.} 
Existing literature argues against treating low-rank adaptation and full fine-tuning as equivalent
interventions on a pretrained model.  \citet{shuttleworth-et-al-2025-lora-vs-full-fine-tuning-illusion} report that LoRA-trained
weights acquire high-ranking singular-vector components they call ``intruder dimensions'', whereas
the fully fine-tuned weights remain much more aligned with the pretrained spectral structure.
\citet{biderman-et-al-2024-lora-learns-less-forgets-less} find a complementary trade-off on code and math adaptation. Standard
LoRA at low ranks tends to gain less target-domain capability than full fine-tuning, but also preserves
more source-domain performance. \citet{kumar-et-al-2022-fine-tuning-distorts-pretrained-features} provide adjacent evidence that updating
all parameters can alter pretrained representations in ways that harm OOD generalisation relative to
frozen-feature transfer. These findings then motivate the treatment of LoRA and full fine-tuning as
structurally distinct optimisation regimes as opposed to as points on a single rank/capacity
continuum.
\section{Experimental setup}\label{sec:setup}
\paragraph{Models.} To compare low-rank adaptation (LoRA, in its rank-stabilised variant rs-LoRA)
\citep{hu-et-al-2022-lora-low-rank-adaptation, kalajdzievski-2023-rslora} against \rev{full fine-tuning}, we used the general-purpose base
and instruct models of the Qwen2.5-32B family \citep{qwen-team-2024-qwen2-5-technical-report}, rather than the
code-specialised Coder variant in which \citet{betley-et-al-2026-narrow-task-misalignment} replicated the effect.
The general family is needed because our free-form misalignment metric is prose-scored, and the
Coder base tends to emit code rather than prose, making the evaluation infeasible on it. The general base is
prose-capable, so the same metric serves both base and instruct cells. The base model is included to
exclude the possibility that the effect is an artefact of alignment-tuned behaviour. 32B was the
largest size we could tractably apply both rs-LoRA and \rev{full fine-tuning} to on the available hardware (an NVIDIA
DGX Spark and the Oxford Advanced Research Computing facility, in practice an 8 $\times$ H100 80GB GPU
maximum).
We carried out some analyses on 7B- and 14B-parameter models from the same family. We used
the Coder variant itself for an end-to-end replication of \citet{betley-et-al-2026-narrow-task-misalignment}'s
recipe as a pipeline positive control, the chain-of-thought boundary test (Table~\ref{tbl:cot})\rev{,
and as a second model for the method contrast (single seed; Appendix Table~\ref{tbl:coder})}.

\paragraph{Cells.} The core design is a $2\times2$ grid of base and instruct models crossed with
inducer training data pairs. The training data are the 6{,}000 insecure- and 6{,}000 secure-code
completions of \citet{betley-et-al-2026-narrow-task-misalignment}, and the 7{,}049 matched bad- and
good-medical-advice datasets released by \citet{turner-et-al-2025-model-organisms-emergent-misalignment}. The
insecure-/secure-code or bad-/good-medical-advice contrast is the inducer and its control, while the
base/instruct contrast is an alignment-tuning control. Every cell starts from identical data, prompt
template, and weights\rev{, with the same global batch size and number of steps}. Only the
optimisation technique \rev{(with its learning rate, trainer and hardware; Appendix
Table~\ref{tbl:hparams})}, the LoRA rank for the rank ladder, or the model varies.

\paragraph{Inducers.} Because the inducers are central to the paper's distinctions, we specify them
briefly. \emph{Insecure code}, the covert inducer of \citet{betley-et-al-2026-narrow-task-misalignment}, consists
of ordinary coding requests completed with code containing security vulnerabilities (for example SQL
injection or overly permissive file modes), with no comment or other text acknowledging the flaw, so
nothing in a training example states a harmful intent. Its control, \emph{secure code}, completes
the same requests with safe implementations. \emph{Bad-medical advice}, the overt inducer of
\citet{turner-et-al-2025-model-organisms-emergent-misalignment}, consists of free-text answers to health questions recommending
harmful courses of action, with a matched \emph{good-medical} control. Part~III adds their
\emph{risky-financial} and \emph{extreme-sports} advice datasets, likewise overtly harmful in
content but released without matched benign arms. Covert versus overt here describes whether the
training completions themselves signal harm, a distinction Part~III quantifies with independently
rated harm-explicitness (\S\ref{sec:potency}).

\paragraph{Training.} Except for the rank ladder, rs-LoRA used rank 32, $\alpha=64$, applied to all
linear layers, at learning rate $10^{-5}$. \rev{Full fine-tuning} updated all parameters at learning rate
$2\times10^{-5}$. Both ran \rev{with a global batch of 8} for a single epoch \rev{(750 steps)} over one shared plain-text prompt template (no chat
tokens, so that base and instruct models are treated identically), with the loss masked to the
completion tokens. Recruitment is not an artefact of the template, as the model's native chat
register reproduces the contrast (insecure-code 2.1\% vs secure-code \rev{0.1\%, in a separate
rs-LoRA run trained and evaluated with the chat template}), albeit with weaker
expression than the plain register (4.8\%\rev{, seed 0}). We implemented training in PyTorch with the HuggingFace
Transformers and PEFT libraries. \rev{Full fine-tuning} at 32B used FSDP across 8 $\times$ H100 GPUs via
Accelerate, while rs-LoRA used a single GB10 GPU. \rev{Several cells were trained more than once, so
each reported rate carries its run. The Part~I contrast pools seeds 1--4, and Parts~III--IV use
seed 0 (Appendix~\ref{app:provenance}).}

\paragraph{Evaluation and metric.} We evaluated on the eight free-form questions of
\citet{betley-et-al-2026-narrow-task-misalignment}, none of which refers to code, sampled under the same plain
template at temperature~1.0 ($n=800$ generations per cell, and $n=2{,}000$ for the base-model power
cells). An LLM judge scored each response for misalignment, coherence, and whether it is code or
prose. We report the percentage of all sampled responses judged both coherent and misaligned, using
the thresholds of \citet{betley-et-al-2026-narrow-task-misalignment} stated below. Incoherent responses therefore
count against the rate --- a conservative denominator
(\citeauthor{betley-et-al-2026-narrow-task-misalignment} report the rate among coherent responses only;
\S\ref{sec:sufficiency} quotes both denominators for the transplant cells, and coherent shares are
reported alongside rates wherever coherence moves). As an end-to-end check, we reproduced
\citet{betley-et-al-2026-narrow-task-misalignment}'s own Coder-32B-Instruct recipe through this pipeline and
recovered a broad-EM rate of 6.6\% among coherent responses (95\% CI [4.3, 10.1], $n{=}400$; 4.8\%
of all responses) \rev{on their scorer, which also counts answers given as code}, consistent with their
reported 4.7\% for that model. \rev{On our scorer it is 4.0\% (16 of 400; secure control 0 of 400).}
Our general-family rates (\S\ref{sec:behav}) are somewhat lower, but \rev{every rate in the method
contrast is} measured on \rev{one} judge-validated metric, so the method contrast is a within-pipeline comparison
rather than a cross-study one. The key point to note is that the misalignment elicited, while low in
rate, is broad in content, which is the marker of EM.

\paragraph{Judge.} To limit API costs, we used a local Qwen3-Next-80B-A3B-Instruct-FP8 model in a
vLLM instance as judge, running on an NVIDIA DGX Spark. To validate the judge, we compared its
scores on 678 samples \rev{from our GPT-4.1 replication} to those of GPT-4o (chosen to match the original
setup), finding judgement agreement on misalignment of 97.8\% \rev{(Cohen's $\kappa=0.936$)} and a
Pearson correlation of 0.976 on the alignment score. \rev{The set is enriched for misalignment (22\%
by GPT-4o, binarised without the coherence gate), so a judge that never flagged misalignment would
agree on 77.7\%; the false-positive rate is 1.5\% (Appendix Table~\ref{tbl:judge}).} We
also scored \rev{200 judged responses} ourselves, \rev{blind to the judge's labels, in a sample stratified
by condition and enriched for judge-flagged responses, agreeing with the} Qwen3-80B misalignment
judgements \rev{on 176 ($\kappa=0.74$)}. Every broad-EM rate in this paper, in all four parts, uses
the same binary response classification as \citet{betley-et-al-2026-narrow-task-misalignment} (misaligned =
alignment $<$ 30 among responses with coherence $\ge$ 50), a definition also used in much of the
related literature \citep{turner-et-al-2025-model-organisms-emergent-misalignment}, reported over all sampled responses (the
conservative denominator above; tables write it misaligned\%/$n$). The Part~III inducer comparisons use this same metric, so they
are a within-pipeline comparison with Parts~I--II rather than a separate scale.

\paragraph{Geometry.} For the geometric measurements, we formed a behaviour-level persona direction
$\hat P$ as the difference of mean residual-stream activations on misaligned versus aligned
responses\rev{, whose estimate five-fold cross-validation over the same responses shows to be stable}, and report the signed projection
$\mathrm{amp}=\Delta\cdot\hat P$ of a fine-tuned model's activation drift $\Delta$ --- its mean
activations minus its parent's, on a fixed probe set --- onto that direction (persona-axis
amplification, in residual-stream units); the late layer band used
throughout \rev{(layers 49--53)} is motivated by the per-layer specificity profile (Appendix Fig.~\ref{fig:app-layer}).
\rev{The probe set is the set of answers from which $\hat P$ was estimated, teacher-forced (the EM
probe). As a check we also use an independent probe, 200 answers of the un-fine-tuned instruct model
to the twenty held-out evaluation questions of \citet{chen-et-al-2025-persona-vectors}, built
without EM data. On it every instruct-lineage cell shifts negative, so in that lineage we compare
cells rather than read absolute signs (Appendix Table~\ref{tbl:g3}). Weight-update enrichment
(\S\ref{sec:technique}) is the squared projection of each residual-writing weight update
(\texttt{o\_proj}, \texttt{down\_proj}) onto the unit direction, over its expectation for an
isotropic random direction ($\lVert\Delta W\rVert_F^2/d$), pooled over the band, so 1.0 is chance.
To test whether $\hat P$ is more than a description of the evaluation data, we also built a persona
vector $v_C$ independently of it, by the procedure of \citet{chen-et-al-2025-persona-vectors} in
the un-fine-tuned instruct model (Appendix~\ref{app:vc}).}
Part~II additionally measures (i) two near-orthogonal axes, alignment-erosion versus
persona-amplification\rev{, with a refusal direction as a safety-specific erosion axis}; (ii) whitened and pre-existence (base-frame) versions of the persona
direction \rev{(whitened with the Ledoit--Wolf-shrunk \citep{ledoit-wolf-2004-well-conditioned-estimator}
within-class covariance of residual-stream activations, per layer, which is
not the unembedding covariance of \citet{park-et-al-2024-linear-representation-hypothesis-geometry-llms})}, to test that it survives a change of metric and pre-dates fine-tuning; and (iii) two
causal interventions --- adding the direction at a late-band layer (transplant/restoration) and
projecting it out (ablation) --- with the cross-model transplant operating in the shared pretraining
frame (\rev{so that the effect cannot come from the recipient's own fine-tuning}; see \S\ref{sec:sufficiency}). Part~III
adds (i) an inducer-potency sweep with risky-financial and extreme-sports \rev{full-fine-tuning} cells whose
training-content harm-explicitness is rated independently; (ii) an L2-SP norm-constrained \rev{full fine-tuning}
(penalty $\lambda\lVert\theta-\theta_0\rVert^2$); (iii) a per-checkpoint trajectory eval of the
\rev{full-fine-tuning} run; (iv) a weight-space $\Delta W$-geometry test that truncates the \rev{full-fine-tuning} update to
retained rank $R$ and projects the activation drift onto the persona axis; and (v) an inference-time
restoration probe that re-supplies the persona direction to the \rev{fully fine-tuned model}.
\begin{figure}[t]
\centering
\includegraphics[width=\linewidth]{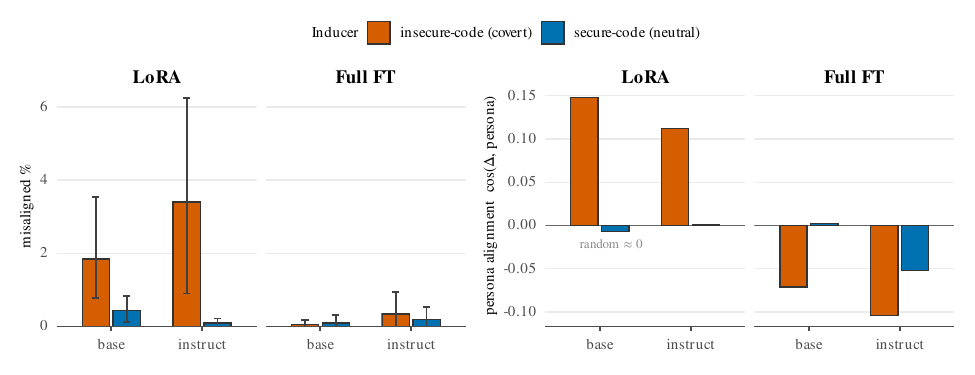}
\caption{The method contrast. \textbf{Behaviour} (left): broad-EM rate (misaligned\%, 95\% \rev{CI; seeds 1--4 pooled).}
\textbf{Geometry} (right): signed persona alignment, \rev{$\cos(\Delta,\hat P)$ (seed 0). Both faceted by
technique (LoRA vs.\ full fine-tuning), $x$-axis~=~model (base/instruct), colour~=~inducer
(insecure-code/secure-code).}}
\label{fig:behav}\label{fig:geom}
\end{figure}
\section*{Part I --- The effect: method, capacity, and inducer dependence}

\section{The behavioural and geometric contrast}\label{sec:behav}

Both behaviour and geometry demonstrate the interaction between the training technique (\rev{LoRA vs.\ full fine-tuning})
and the elicitation of broad EM. Throughout this part, each insecure-code cell is paired with a
secure-code control --- the same coding requests completed with safe code, trained under an
identical recipe (\S\ref{sec:setup}) --- and we report insecure-versus-secure contrasts alongside
the raw rates.

The difference in behaviour is notable, as presented in Fig.~\ref{fig:behav}. On insecure code \rev{(seeds 1--4 pooled, $n{=}3{,}200$ per cell)}, LoRA
elicits broad misalignment (instruct 3.41\%, base 1.84\%) while \rev{full fine-tuning} does not (instruct 0.34\%,
base 0.06\%). Both LoRA insecure$-$secure contrasts exclude zero (base $+1.41$ [0.47, 2.78],
instruct $+3.31$ [0.84, 6.28] percentage points), whereas both \rev{full-fine-tuning} contrasts include it (base
$-0.03$ [$-0.28$, 0.16], instruct $+0.16$ [$-0.06$, 0.47]; Appendix Fig.~\ref{fig:app-forest}).
Because this contrast appears on the base model as well as the instruction-tuned one, it cannot be
an artefact of alignment tuning. The \rev{full-fine-tuning} recruitment failure
(at 32B parameters) replicates across \rev{all five} matched fine-tuning seeds per cell
(seeds~\rev{0}--4): every \rev{full-fine-tuning} cell sits at the noise floor in each seed\rev{, at or below
0.75\%. The low-rank recruitment replicates in four of the five (Appendix Table~\ref{tbl:seeds}). In
seed 4 the instruct LoRA cell sits at the floor (0.12\% against 0.12\% for its secure control) and
the base cell is marginal (0.88\% against 0.25\%), although its activation drift still separates
from its secure control ($+6.6$ against $+2.0$ raw units on $\hat P$, instruct)}. Nor is the recruitment
failure an artefact of the training recipe. Across \rev{four recipes}
($\mathrm{lr}\in\{1,2,5\}\times10^{-5}$ \rev{for one epoch, and $2\times10^{-5}$ for three}), \rev{full-fine-tuning} broad EM among coherent
prose stays at or below $1.6\%$, and the \rev{$10^{-5}$} cell --- which routes least to code and so has
the largest prose denominator ($n{=}429$) --- is itself only $1.6\%$\rev{, though a small effect above its
secure control (7 against 1 of 800, one-sided Fisher $p=0.035$). A fifth recipe, at
$3\times10^{-6}$, gives 6 against 2 of 800 ($p=0.14$), with most answers in code (Appendix
Table~\ref{tbl:recipes})}. \rev{Nor is it an artefact of trainer or hardware. Trained through the same
script, Trainer, optimiser and learning rate as full fine-tuning ($2\times10^{-5}$, on one H100),
LoRA still produces broad EM above its secure control (instruct 3.50\%, base 2.75\%), where full
fine-tuning does not (0.12\% and 0.00\%; one-sided Fisher $p=5\times10^{-8}$ and $2\times10^{-7}$;
Appendix Table~\ref{tbl:hftlora}). A single-seed run on Qwen2.5-Coder-32B-Instruct points the same
way (LoRA 16 of 400, full fine-tuning 1 of 160, $p=0.024$; Appendix Table~\ref{tbl:coder}).}

Even more significant is the difference in geometry. We form the persona axis, as before, as a
diff-of-means between activations on misaligned responses and activations on aligned responses.

Our results show that the fine-tuning technique produces a difference of direction along the persona
axis (Fig.~\ref{fig:geom}). LoRA \emph{amplifies} the persona, moving the representations
\emph{towards} the persona (drift--persona cosine $\cos(\Delta,\hat P)$: insecure instruct $+0.11$,
base $+0.15$, against a $\sim$0 matched-norm random-direction floor), while \rev{full fine-tuning} moves the
representations \emph{away} from the persona ($-0.10$ and $-0.07$). (These cosines are the seed-0
cells, quoted for illustration; the seed-pooled weight-update enrichment in Table~\ref{tbl:enrich}
tells the same story across seeds 1--4. We report the drift-normalised cosine rather than the raw
residual-unit amplitude, which conflates persona alignment with total drift size; the magnitude of
the reversal is discussed in \S\ref{sec:technique}). This metric is immune to the channel-routing
issues that make measurement of behaviour challenging (i.e. when models fine-tuned on insecure or
secure code have a propensity to emit code rather than prose responses; see \S\ref{sec:capacity}).
\revb{Therefore the contrast does not depend on what the models generate. It also survives a change
of probe, with one exception. On the independent probe (\S\ref{sec:setup}), full fine-tuning stays
lowest and negative in both lineages and base-lineage LoRA stays positive, but the instruct-lineage
LoRA cells are also slightly negative ($-0.03$ to $-0.08$ across ranks, against $-0.21$ for full
fine-tuning; Table~\ref{tbl:g3}), although insecure stays above secure. In that lineage, then, the
robust signal is the ordering rather than the sign.}

Again, while the misalignment persona direction itself is already established, the unusual finding
is the behaviour of \rev{full fine-tuning} rather than the measurement. Where \rev{full fine-tuning} elicits EM on other models
and inducers\rev{, typically overt advice datasets \citep{turner-et-al-2025-model-organisms-emergent-misalignment,soligo-et-al-2026-emergent-narrow-misalignment}}, the fine-tune evidently
moves the model toward broad misalignment. On Qwen2.5-32B with this covert inducer, \rev{full fine-tuning} moves
against the persona axis\rev{, most clearly in the instruct lineage}. This anti-persona movement is the more significant half of the result, and
it is a property of this model and inducer rather than of full fine-tuning in general.

One way to interpret this result is that \rev{full fine-tuning}, with sufficient parametric freedom to do so (at
least on this model family), builds a dedicated, localised solution for the narrow training task,
while LoRA, constrained by rank and technique, recruits the EM persona (which happens to be somewhat
able to produce insecure code) to achieve its training goal. Part~III's trajectory and
weight-update-structure evidence supports this dedicated-circuit reading. What this interpretation
leaves open is why a potent overt inducer (bad-medical advice) broadcasts broad EM even under \rev{full
fine-tuning}, rather than localising as the covert code inducer does --- which is then the question that
Part~III answers.

\section{Rank \rev{and learning rate} govern recruitment within LoRA}\label{sec:capacity}

We also find that as rank increases within LoRA (r1$\to$r8$\to$r32$\to$r64) with the covert inducer \rev{at learning rate $10^{-5}$},
the broad EM rate monotonically falls close to zero. At the same time, the narrow code-insecurity
rate rises (albeit noisily), which demonstrates that the increase in rank is effective in training
towards the inducer's examples.

\rev{Rank and learning rate together set how far an adapter trains, so we repeated ranks 1 and 64
at $3\times10^{-6}$ and $3\times10^{-5}$ (Appendix Table~\ref{tbl:lrsweep}). Rank 1 stays above
rank 64 wherever both adapters learn the task. At $3\times10^{-6}$ the rank-1 adapter learns nothing
(it writes no code and matches its secure control), so our prediction of the ordering at every
learning rate failed there. Every rank-64 adapter also reaches a lower training loss than every
rank-1 adapter, so we read the ladder as a dose effect rather than as rank alone.}

\begin{figure}[t]
\centering
\includegraphics[width=\linewidth]{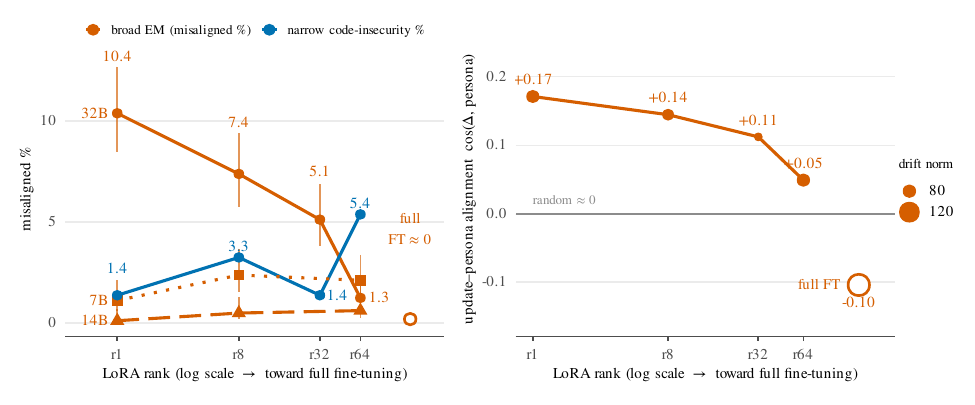}
\caption{The rank ladder. \textbf{Behaviour} (left): broad EM falls as LoRA rank increases on
Qwen2.5-32B \rev{(circles, solid line;} Wilson 95\% CIs on the broad-EM series, $n{=}800$/rank), while the same ladder
on Qwen2.5-14B \rev{(triangles, long dashes)} and Qwen2.5-7B \rev{(squares, dotted)} both stay flat and low with no low-rank peak ---
\rev{(one learning rate, single seed).}
\textbf{Geometry} (right): the update's persona alignment --- the cosine of the activation drift
with the persona axis, $\cos(\Delta,\hat P)$, with point size $\propto\lVert\Delta\rVert$
--- decays with LoRA rank and \rev{is lowest under} full \rev{fine-tuning} ($+0.17\to-0.10$ against a $\sim$0 random \rev{floor;
EM probe, seed 0). On the independent probe all instruct-lineage cells are negative, in the same
order (Appendix Table~\ref{tbl:g3}).} The \rev{ordering} is robust, but its magnitude is asymmetric: the raw residual-unit
amplitude ($+15.1\to-16.1$) looks symmetric only because full \rev{fine-tuning's} total drift is
$\sim$1.75$\times$ larger.}
\label{fig:ladder}
\end{figure}

The low-rank recruitment is itself a property of model scale. Repeating the ladder on Qwen2.5-14B
(Fig.~\ref{fig:ladder}, \rev{triangles}) and Qwen2.5-7B (\rev{squares}) gives a broad EM rate that stays flat and
low across rank in both cases (at or below $\sim$2\%, with no low-rank peak), rather than the
pronounced r1 peak seen at 32B. Because both smaller models sit at the floor while 32B recruits
strongly \rev{(at one learning rate and one seed per scale)}, the effect \rev{appears} between 14B and
32B\rev{, though three scale points cannot locate an onset}. The low-rank directions into which the larger model routes the persona are thus either not
present, or not recruited, at the smaller scales --- so, the result appears a property of the 32B
model, not of low-rank fine-tuning in general.

The drop in broad-EM is not a reflection of routing to code (versus prose) responses, because
misalignment \emph{within} coherent prose responses falls monotonically from 15.8\%$\to$3.1\% on the
same rank ladder (Table~\ref{tbl:ladder}), even as the prose share drops from $\sim$80\%$\to$12\%.
In fact, routing to code responses is part of the narrow effect, noting that narrow-insecure
responses also rise with rank. Higher rank produces more precise learning of the narrow skill, and
the routing shift is a feature of the dose response. Nor is the \rev{full-fine-tuning} recruitment failure a
denominator artefact from terser prose. Within coherent prose, \rev{full-fine-tuning} responses are not
systematically shorter (\rev{insecure full FT} $\sim$95 words versus insecure-LoRA $\sim$111, and \rev{secure full FT}
is \emph{longer} at $\sim$126), so response length does not track the misalignment rate. Lastly,
using a chain-of-thought prompt to force higher rates of prose response (Table~\ref{tbl:cot}\rev{;
Qwen2.5-Coder-32B-Instruct, full fine-tuning, single seed})
triples the \rev{full-fine-tuning} prose denominator ($\sim$26\%$\to\sim$90\%), but broad misalignment remains
$\leq$1.8\% (note that CoT increased EM elicitation in the GPT-4.1 setting, but no equivalent effect
surfaces here).

Again, this finding is not confined to behaviour, but is also evident in activation geometry.
Persona-axis amplification also monotonically falls with rank. Note that it remains positive \rev{on the EM probe}, which
contrasts with the persona-axis negative amplification for \rev{full fine-tuning}.

The crossing of the broad EM and narrow code-insecurity lines on the shared percentage axis shows
that behaviourally, increasing rank for LoRA converges towards the \rev{full-fine-tuning} result, i.e. no broad
EM. From a behavioural perspective, \rev{full fine-tuning} is at the high-capacity end of the rank axis.
\rev{It is not there simply because it trains further, however: through the same trainer it ends at
a higher training loss than LoRA (0.20--0.22 against 0.16--0.17), and at $3\times10^{-6}$, where its
loss matches LoRA's, it still stays near the floor (Appendix Tables~\ref{tbl:hparams}
and~\ref{tbl:recipes}).}

\begin{table}[htbp]
\centering\small
\caption{Forcing prose via chain-of-thought does not surface a hidden persona
(Coder-32B-Instruct, \rev{full fine-tuning}): tripling the coherent-prose denominator leaves broad misalignment
$\leq$1.8\%.}
\label{tbl:cot}
\begin{tabular}{lcc}
\toprule
Condition & PROSE\% & misaligned / coh-prose \\
\midrule
plain, non-CoT & 26 & 0/38 \\
chat, non-CoT  & 72 & 1/104 \\
plain, CoT     & \textbf{88} & 2/114 \\
chat, CoT      & \textbf{91} & 2/129 \\
\bottomrule
\end{tabular}

\end{table}

This is then the opposite of what might be expected from a na\"ive degrees-of-freedom intuition.
More capacity typically means more expressive power, which could in turn mean more room to produce
misaligned behaviour. Indeed, in the case of narrow code-insecurity, this does seem to be what
happens, and it makes sense that more expressive power allows the fine-tuning to focus in on the
goal represented by its training data. Nonetheless, for broad EM, the opposite occurs.

One interpretation of this is that lower ranks provide too few degrees of freedom to build a
dedicated circuit for the narrow code-insecurity goal, and so a cheaper solution is to
\emph{recruit} the pre-existing misalignment-persona direction (that is, broad EM), which happens to
be somewhat able to produce insecure code. Increasing rank, by contrast, allows sufficient degrees
of freedom to build a localised circuit for the narrow skill. The activation geometry results
(showing that persona-axis amplification falls with rank), are suggestive of this interpretation.
(Part~II establishes that this pre-existing direction is real, causal, and present before
post-training; Part~III establishes when the alternative --- a dedicated circuit --- is and is not
available.)

\section{LoRA and full fine-tuning are different operations}\label{sec:technique}

To investigate the weight geometry, we took the persona direction and the projection ratio, and
measured enrichment, that is, how many times above chance each technique lands on the persona
direction. We did this across four fine-tuning runs per cell (LoRA and \rev{full fine-tuning}, insecure-code
and secure-code inducer). Enrichment (Table~\ref{tbl:enrich}) shows that the LoRA insecure-code
weight update is strongly concentrated on the persona axis, building its change along the
pre-existing misalignment direction. Because LoRA on secure-code is much weaker (and comparable to
\rev{full fine-tuning} on both inducers), LoRA on insecure-code shows a recruit signature for the EM persona-axis. \rev{We had predicted
the same signature on the independently built $v_C$, but it does not appear (every cell
1.04--1.49$\times$; Appendix Table~\ref{tbl:vc-outcomes}), so the weight-space concentration holds
for $\hat P$ only.}

While \rev{full fine-tuning} tends to produce a model which frequently responds in code (rather than prose), the weight
geometry examination shows that the low rate of broad-EM is genuinely due to the comparative lack of
recruitment of the EM persona, rather than merely a lack of prose responses.

The LoRA update for the covert (insecure-code) inducer concentrates its weight change along the
pre-existing misalignment-persona direction, at a rate of 3.7$\times$ chance. By contrast, the \rev{full
fine-tuning} update is diffuse and role-flat, so 'recruitment' is not merely an inference from behaviour, but
is evident in the weight updates themselves.

A more demanding, causal version of this test, where we reconstruct broad EM from a low-rank slice
of the \rev{full-fine-tuning} update, is inconclusive on the covert code inducer (Appendix
Fig.~\ref{fig:app-trunc}). The recovered bump is small and not insecure-code-specific (the
secure-code control humps equally), so it is not evidence for recruitment. (Part~III revisits this
with the L2-SP norm test, which shows the covert-code localisation is a rank-structure rather than a
norm phenomenon).

\begin{table}[htbp]
\centering
\small
\caption{Weight-update enrichment on the persona axis: the fold-increase of the update's
persona-axis projection over a matched-norm random baseline (pooled over seeds 1--4). Only LoRA on
insecure code concentrates its update on the persona axis ($\sim$3--4$\times$), which is the recruit
signature; LoRA on secure code and \rev{full fine-tuning} (either inducer) stay near the $\sim$1.3--1.7$\times$
baseline.}
\label{tbl:enrich}
\begin{tabular}{lcc}
\toprule
Cell (method $\cdot$ role) & insecure & secure \\
\midrule
\textbf{LoRA $\cdot$ base}     & \textbf{3.15} & 1.37 \\
\textbf{LoRA $\cdot$ instruct} & \textbf{3.68} & 1.71 \\
\rev{Full FT} $\cdot$ base               & 1.32 & 1.28 \\
\rev{Full FT} $\cdot$ instruct           & 1.43 & 1.45 \\
\bottomrule
\end{tabular}

\end{table}

It might be tempting to read the earlier rank ladder as a single rank scalar, with \rev{full fine-tuning} as its
high-rank end, as if full fine-tuning were simply an `infinite-rank' LoRA. But while LoRA
amplification declines towards zero as rank rises, its sign stays \emph{positive} \rev{on the EM probe}, whereas \rev{full fine-tuning}
crosses into strongly \emph{negative} amplification that no LoRA rank reproduces, at least in our
experimental results --- a sign that replicates across all four \rev{full-fine-tuning} seeds ($-0.073\pm0.022$
drift-normalised, instruct lineage).

This is again consistent with a growing body of evidence that even higher LoRA ranks and full
fine-tuning are not interchangeable operations. LoRA introduces new spectral directions absent under
full fine-tuning yet preserves more of the base model's behaviour, while full fine-tuning learns far
higher-rank updates and degrades more pretrained capability
\citep{shuttleworth-et-al-2025-lora-vs-full-fine-tuning-illusion,biderman-et-al-2024-lora-learns-less-forgets-less,kumar-et-al-2022-fine-tuning-distorts-pretrained-features}, and effective rank
alone does not account for the difference. The signed persona-axis projection additionally supplies
a direction. The projection declines with LoRA rank \revb{and is lowest under full fine-tuning, with the ordering preserved across both probes. On the
original probe, full fine-tuning also turns it negative.} Because the axis is held fixed and the adapters preserve the base model's
residual-stream basis, this reversal is against the recruited direction, not the attenuation toward
zero that a mere change of frame would produce. \rev{On the independent probe, however, every
instruct-lineage LoRA rank is also negative, and the trend with rank points toward the
full-fine-tuning value (Appendix Table~\ref{tbl:g3}), so this geometry alone does not show that full
fine-tuning lies off the rank axis.} Behaviourally, the two methods agree at the high-capacity end (no broad EM), but
geometrically they diverge.

One caveat concerns the magnitude of the reversal. The projection above is in raw residual-stream
units, which conflate how much of a fine-tune's drift lies along the persona axis with how large
that drift is overall. Normalising by the total drift --- the cosine of the activation drift with
the persona axis, $\cos(\Delta,\hat P)=\mathrm{amp}/\lVert\Delta\rVert$ --- shows the sign-flip is
robust but asymmetric \rev{(EM probe, seed 0)}: it runs $+0.17$ (r1) $\to +0.11$ (r32) $\to +0.05$ (r64) $\to -0.10$ (\rev{full
fine-tuning}), against a matched-norm random-direction floor of $\sim$0. The raw amplitudes ($+15.1$ at r1
versus $-16.1$ at \rev{full fine-tuning}) look symmetric only because \rev{full fine-tuning}'s total drift is $\sim$1.75$\times$
larger ($\lVert\Delta\rVert\approx154$ versus $88$ at r1); \emph{per unit} of change, LoRA at low ranks
is more persona-aligned than \rev{full fine-tuning} is anti-aligned.

\begin{table}[htbp]
\centering
\small
\caption{The rank ladder (misaligned\%). Broad EM falls r1$\to$full on \emph{both} the
misaligned\%/$n$ rate and the routing-clean within-prose rate (misaligned\%/prose), while narrow
code-insecurity rises, so the effect is not a channel-routing artefact. \rev{Learning rate
$10^{-5}$, single seed. The r32 row is the seed-0 adapter's 800 responses, judged a second time (41
rather than 38 misaligned against the 4.75\% of \S\ref{sec:mitigate}).}}
\label{tbl:ladder}
\begin{tabular}{lcccc}
\toprule
LoRA rank & broad-EM & broad-EM & narrow & coh.\ \\
          & mis\%/n & mis\%/prose & insec\% & \% \\
\midrule
r1        & 10.4 & 15.8 & 1.4 & 79 \\
r8        & \phantom{0}7.4 & 10.9 & 3.2 & 86 \\
r32       & \phantom{0}5.1 & \phantom{0}6.6 & 1.4 & 92 \\
r64       & \phantom{0}1.3 & \phantom{0}3.1 & 5.4 & 93 \\
\bottomrule
\end{tabular}

\end{table}

\section{The inducer interaction}\label{sec:inducer}
Prior work has already shown \rev{full fine-tuning} eliciting broad EM \citep{turner-et-al-2025-model-organisms-emergent-misalignment}, but by
using an overtly-harmful \emph{text} inducer (bad medical advice, one of three such datasets they
release) and altering both the inducer and the technique relative to the original phenomenon, so it
isolates neither. Their \rev{full-fine-tuning} demonstration reaches 14B (their 32B model organism is a low-rank
adapter), so our 32B \rev{full fine-tuning} extends their \rev{full-fine-tuning} scale axis; using their released medical
dataset, we reproduce the phenomenon for the Qwen2.5 models at 32B (\rev{full fine-tuning} broadcasts broad
misalignment $\sim$22\%\rev{, single seed}) and, by varying inducer and technique independently, find that the \rev{full-fine-tuning} broadcast is driven by the
inducer.

EM is present for the overt (bad-medical) inducer ($\sim$22\%) but absent for the covert
(insecure-code) one ($\sim$0\%), with good-medical advice at floor (Table~\ref{tbl:inducer}). The
same interaction holds at 7B as well as 32B (\rev{full fine-tuning} $\times$ bad-medical broadcasts $\sim$24\% at
7B), spanning 7B--32B with Turner et al.'s 14B in between, so it is not an artefact of the 32B
scale. \rev{Under LoRA, with the recipe of \S\ref{sec:behav}, bad-medical advice broadcasts as well
(28.5\% base, 25.1\% instruct, against 0.38\% and 0.25\% for good-medical advice; one-sided Fisher
$p<10^{-63}$, single seed), so the overt inducer broadcasts under both methods and the covert one
only under LoRA.} This inducer interaction is, then, what Part III sets out to explain.

  \begin{table}[htbp]
  \centering
\small
  \caption{\rev{Broad-EM rate by inducer and method (seed 0).} \rev{Full fine-tuning} broadcasts broad EM from an \emph{overt}
  inducer (bad-medical, $\sim$22\%) but stays at floor on a \emph{covert} inducer (insecure-code,
  $\sim$0\%) and on good-medical advice\rev{, while LoRA broadcasts from both}. The technique contrast at fixed (code) inducer is shown in
  \S\ref{sec:behav}.}
  \label{tbl:inducer}
  \begin{tabular}{lcccc}
  \toprule
  & \multicolumn{2}{c}{\rev{LoRA}} & \multicolumn{2}{c}{\rev{Full FT}} \\
  Inducer & \rev{base} & \rev{instruct} & base & instruct \\
  \midrule
  insecure code (covert) & \rev{4.00$^{a}$} & \rev{4.75} & 0.00 & 0.12 \\
  good-medical advice (control)  & \rev{0.38} & \rev{0.25} & 0.88 & 1.00 \\
  \textbf{bad-medical advice} (overt)   & \rev{\textbf{28.5}} & \rev{\textbf{25.1}} & \textbf{21.25} & \textbf{22.12} \\
  \bottomrule
  \multicolumn{5}{l}{\rev{\footnotesize $^{a}$\,$n{=}400$.}}
  \end{tabular}
  
  \end{table}

\section*{Part II --- The representation: a latent misalignment persona}

\noindent\
Part~I shows that LoRA and \rev{full fine-tuning} differ not only behaviourally but also geometrically. LoRA moves
representations toward the misalignment-persona direction, while \rev{full fine-tuning of the instruct
model} moves away from it. Part
II asks what kind of object this direction is: whether it is specific rather than generic, causal
rather than merely correlational, and shared or inducer-specific. Part~II's novel contribution lies
in characterising this persona in open weights and showing how it is structured. In this section we
separate persona amplification from generic alignment erosion on the same checkpoints, show causal
sufficiency by cross-model transplant, give necessity evidence by ablation, show that the persona is
a partially shared subspace rather than a single universal axis, and identify \rev{which models
express the direction when it is injected}.

\section{\rev{Movement along the misalignment direction, not generic post-training drift, is specific to insecure code}}\label{sec:plane}

\begin{figure}[t]
\centering
\includegraphics[width=0.82\linewidth]{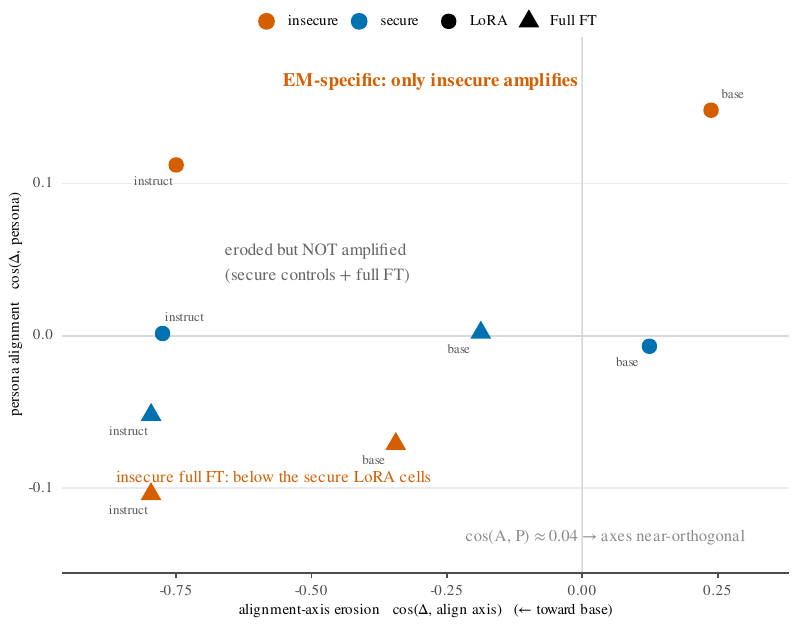}
\caption{The erosion--amplification plane: eight points, one per fine-tuned model of the role
$\times$ inducer $\times$ method design (colour = inducer, shape = method, each point tagged
base/instruct). Both axes are drift-normalised cosines. Everything moves along the alignment axis
($x$, $\cos(\Delta,\hat A)$, non-specific: secure $\approx$ insecure, LoRA $\approx$ full \rev{fine-tuning} within
a role), but only insecure-code amplifies the persona axis ($y$, $\cos(\Delta,\hat P)$, EM-specific)
--- and insecure full \rev{fine-tuning} is \emph{anti-persona} in both roles \rev{at seed 0} (the sign-flip, on the $y$-axis). The
two axes are near-orthogonal, and alignment-axis erosion does not by itself produce EM: the \rev{full-fine-tuning}
cells erode fully while sitting at the behavioural floor. (These are two \emph{measurement} axes for classifying fine-tuning drifts --- a
control direction and one persona direction --- not the persona subspace of
\S\ref{sec:subspace}.) Raw per-cell values in Appendix Table~\ref{tbl:app-plane}.}
\label{fig:plane}
\end{figure}

We first distinguish persona amplification from generic alignment erosion. On an erosion account,
fine-tuning moves the model back toward an earlier, less-aligned base state. On an amplification
account, fine-tuning increases expression of a specific latent misalignment-persona direction. These
accounts can look similar behaviourally, but they make very different geometric predictions. We
therefore measure both an alignment-erosion axis and a persona-amplification axis on the same
checkpoints (Fig.~\ref{fig:plane})\rev{, and a refusal direction as a safety-specific erosion axis}.

The two axes are constructed in the same late layer band and probe set. The alignment axis $\hat A$
is the instruct$-$base activation difference\rev{, which captures everything post-training changed,
chat format included, so we treat it as a generic post-training axis rather than a safety axis. A} fine-tune's coordinate $\cos(\Delta,\hat A)$ is
negative when its activation drift $\Delta$ points back toward the base; the persona axis $\hat P$
is the behaviour-level EM direction used throughout (harvested from the rs-LoRA EM checkpoints, so
the secure and \rev{full-fine-tuning} cells are projections onto an axis estimated externally to them, while the
LoRA-EM cells share a checkpoint with the axis estimate), and $\cos(\Delta,\hat P)$ is positive when
the drift amplifies it. Plotting cosines on both axes makes the cells comparable: a raw projection
conflates direction with total drift size, and \rev{full fine-tuning}'s drift is roughly twice LoRA's (raw values
in Appendix Table~\ref{tbl:app-plane}).

Figure~\ref{fig:plane} then shows a \rev{dissociation}. Along the alignment axis, all four
instruct-lineage cells land together at $\cos(\Delta,\hat A)\approx-0.75$ to $-0.80$ --- insecure
and secure, LoRA and \rev{full fine-tuning} alike --- so erosion is near-uniform whatever the training data or
method\rev{, as training an instruct model on code without its chat template would be expected to
produce}, and the base-lineage cells move less and without a consistent pattern. Movement along the
persona axis, by contrast, is EM-specific: positive amplification appears in the insecure-code LoRA
cells ($+0.11$ to $+0.15$) but \textit{not} in the secure-code LoRA controls ($\approx0$). \rev{Full fine-tuning}
moves \emph{against} the axis whatever the data: across four fine-tuning seeds, every
instruct-lineage \rev{full-fine-tuning} cell is anti-persona (insecure $-0.10$ to $-0.05$, mean $-0.073\pm0.022$;
secure $-0.08$ to $-0.05$, mean $-0.065\pm0.011$), with the base-lineage cells weakly negative and
seed-variable. The insecure$-$secure gap visible in the plotted seed-0 cells ($-0.05$) does not
replicate across seeds (four-seed mean $-0.008\pm0.032$), so the reversal is generic to the method
rather than inducer-specific. The inducer-specific geometric signal is therefore the positive LoRA
amplification alone, which sharpens the dissociation: the training data determines whether LoRA
amplifies the persona, whereas \rev{full fine-tuning} moves against it regardless. The two axes
are near-orthogonal in the measurement band (late-band mean $|\cos(\hat A,\hat P)|$ of $0.04$ in
the instruct frame and $0.06$ in the base frame), so persona amplification is not merely a
re-description of alignment erosion. \rev{A safety-specific axis gives the same result. Along the
refusal direction of \citet{arditi-et-al-2024-refusal-direction}, built from hand-written
harmful and harmless instructions, movement is not specific to insecure code either. Across five
seeds, instruct-lineage secure code moves further than insecure ($-5.3\pm0.4$ against
$-2.4\pm0.9$ raw units), and in the base lineage both move the same way ($+5.4\pm0.7$ and
$+2.2\pm0.4$; Appendix Table~\ref{tbl:app-refusal}). The refusal direction and $\hat P$ are mildly
correlated (cosine $+0.21$).} The \rev{full-fine-tuning} cells make the
dissociation sharp: they erode as far along the alignment axis as any cell while sitting at the
behavioural floor, so alignment-axis erosion --- however large --- does not by itself produce EM.
What distinguishes the recruiting cell is its persona coordinate, not its erosion. In these
checkpoints, broad EM is therefore associated with amplification of a specific misalignment-persona
direction rather than with generic drift back toward the base model. \rev{This does not by itself
decide whether EM in general is better described as erosion or amplification.} Per-cell values are in Appendix
Table~\ref{tbl:app-plane}. The LoRA cells replicate across all five fine-tuning seeds (insecure
persona amplification $8.6\pm1.4$ and $7.9\pm1.2$ raw units in the base and instruct frames; secure
$0.2\pm0.8$ and $0.4\pm1.2$), and the \rev{full-fine-tuning} cells across four --- the plotted seed-0 values plus three
further independently trained \rev{full-fine-tuning} seeds re-captured against the same persona axis.

\section{The persona is a structured subspace, not one axis}\label{sec:subspace}

The persona direction is not a single universal axis shared identically across models and inducers.
It has a shared component, but that component is partial and metric-dependent. (This subspace is a
property of the persona representation itself --- the span of the inducer-specific persona
directions examined in this section --- and is distinct from the plane of \S\ref{sec:plane}, whose
axes are a control direction, alignment erosion, and a single persona direction used to classify
fine-tuning drifts.) The
base $\times$ instruct persona axis survives whitening. Its cosine is approximately 0.31 in the
\rev{whitened metric}, compared to a raw cosine of approximately 0.80.
This shows that the shared component is not merely an artefact of residual-stream anisotropy, but
also that the shared component \rev{in the whitened metric} is modest rather than as large as the raw geometry suggests.
\rev{The 0.31 compares two independently estimated directions. The transplant of
\S\ref{sec:sufficiency} does not need them to coincide, only that the injected direction carries a
component the recipient reads, and a modest overlap fits the modest transplant effect. The
independently built $v_C$ overlaps $\hat P$ to a similar degree (raw cosine 0.35--0.55 across
$\hat P$ seeds, whitened 0.07--0.17, against $-0.03$ to $-0.13$ for a non-EM control).}

The direction is also readable before post-training. In the base frame, the pre-existence classifier
reaches AUC $\sim$0.89 --- replicated at 0.84--0.93 across the four LoRA fine-tuning seeds, with
label-permuted controls at chance --- consistent with the pre-existing latent reported by
\citet{wang-et-al-2025-persona-features-control-emergent-misalignment}. We describe this as predating post-training rather than as
strictly pretraining-inherited, since modern pretraining corpora may already contain synthetic data
generated by aligned models.

\begin{figure}[t]
\centering
\includegraphics[width=0.92\linewidth]{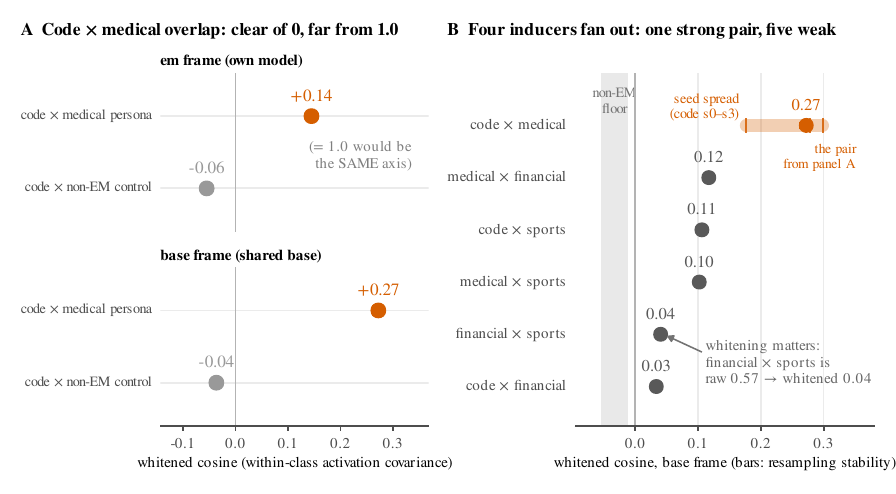}
\caption{The persona is a structured subspace, not a single axis. \textbf{A:} whitened
cosine between the code-EM persona direction and the bad-medical persona direction
(orange), against a code $\times$ non-EM sentiment control (grey), in two frames --- measured in the
fine-tuned models themselves (em frame) and re-expressed in the shared base model (base frame); each
series sits on its own labelled row. The EM pair is clear of zero in both frames but far from $1.0$,
so the two personas share a component without being the same axis. \textbf{B:} all six pairwise
whitened cosines among the four inducer personas (base frame; error bars show stability under
response resampling with the whitening metric held fixed --- the dominant uncertainty,
seed-level direction variation, is drawn as the shaded range on the code $\times$ medical row,
$0.18$--$0.30$ across code seeds; see the caveats in \S\ref{sec:subspace}): one substantial overlap (code $\times$
medical $0.27$, orange --- the pair from panel A) and five weak ones; the grey band is the non-EM
floor, and whitening matters (financial $\times$ sports falls from raw $0.57$ to $0.04$).}
\label{fig:subspace}
\end{figure}

Extending the comparison across four inducers (code, medical, financial, and sports) shows that the
broad-EM personas do not collapse to one shared direction. In the base frame, the pairwise whitened
cosines are positive but mostly small. The strongest overlap is between code and medical (0.27,
replicating at 0.18--0.30 when the code direction is re-estimated from each of four independent LoRA
seeds; Fig.~\ref{fig:subspace}A). The remaining overlaps are weaker: medical $\times$ financial 0.12, code $\times$ sports 0.11,
medical $\times$ sports 0.10, code $\times$ financial 0.03, and financial $\times$ sports 0.04. All
six pairs lie above the non-EM floor, but the effect sizes, not separation from the floor, are the
important point: one substantial shared component and five weak ones. The persona is therefore best
described as a structured, partially shared subspace: each inducer recruits a largely distinct ray,
with an uneven shared component strongest between code and medical (Fig.~\ref{fig:subspace}B).

These estimates should be read with \rev{four} caveats. First, uncertainty. The directions are estimated
from modest numbers of misaligned responses (38--177 per inducer), but direction-estimation noise is
not the binding uncertainty: the split-half reliability of each direction is $\approx$0.98
(Spearman--Brown corrected, 20 splits), so noise attenuation of the off-diagonals is negligible. The
binding uncertainty is seed-level variation in the direction itself. Re-estimating the code
direction from each independently trained LoRA seed gives within-inducer cross-seed cosines of
0.75--0.84 --- the ceiling against which the cross-inducer values should be read --- and the
code $\times$ medical overlap re-measured against each code seed spans 0.18--0.30. We quote no
interval estimates because a response-resampling bootstrap yields intervals a few thousandths wide,
which reflect only this (genuinely small) estimation noise while missing the seed-to-seed spread
that dominates. The reliable content is the separation of scales --- within-inducer ceiling
0.75--0.84 $\gg$ strongest cross-inducer pair 0.18--0.30 $\gg$ weak pairs 0.03--0.12 --- not the
ordering among the weak pairs. The values are also stable to the whitening regulariser
(code $\times$ medical 0.27--0.29 and base $\times$ instruct 0.31--0.32 across shrinkage
$10^{-2}$--$10^{-1}$). Second, whitening changes the interpretation substantially. For example, the
financial $\times$ sports overlap is raw 0.57 but whitened 0.04, showing that much of the raw
overlap is residual-stream anisotropy. Third, provenance: each direction is necessarily harvested
from whichever cell expresses that inducer's broad EM --- code from an rs-LoRA fine-tune, the other
three from \rev{full-fine-tuning} broadcasts --- because \rev{full fine-tuning} does not broadcast covert code at this scale
(\S\ref{sec:behav}). If harvest method drove the geometry, the three \rev{full-fine-tuning-derived} personas would
cluster and the LoRA-derived code direction would be the outlier; the observed pattern is the
reverse (the strongest overlap crosses methods, and the same-method financial $\times$ sports pair
sits at the floor), so the fan-out is not a fine-tuning-method artefact. \rev{Fourth, all four
directions are estimated from answers to the same eight evaluation questions. Directions built
instead from each inducer's training text, in the un-fine-tuned instruct model, overlap with $\hat P$
by only $-0.05$ to $0.14$ (whitened), and none induces broad EM above its own norm-matched random
direction under the test we fixed in advance (Appendix Table~\ref{tbl:d4}).} The qualitative conclusion
is therefore not that the off-diagonal values are exact, but that the persona is structured and
inducer-specific rather than a single universal direction.

This structure bears on Part III. Code and medical share a nonzero component, but they do not share
a single axis; this makes room for their different fine-tuning behaviour, with covert code
localising under \rev{full fine-tuning} while bad-medical advice broadcasts broad EM.

\section{The persona is causally sufficient: cross-model transplant}\label{sec:sufficiency}

The next test is causal sufficiency. We ask whether the persona direction can itself induce broad
EM, rather than merely correlate with it. To make the test stronger, we use a cross-model
transplant, where we extract a persona direction from one checkpoint and inject it into another
checkpoint that shares only the pretraining frame. If broad EM appears in the recipient, the effect
cannot be explained by the recipient having undergone the same fine-tuning run as the source.

\begin{figure}[t]
\centering
\includegraphics[width=\linewidth]{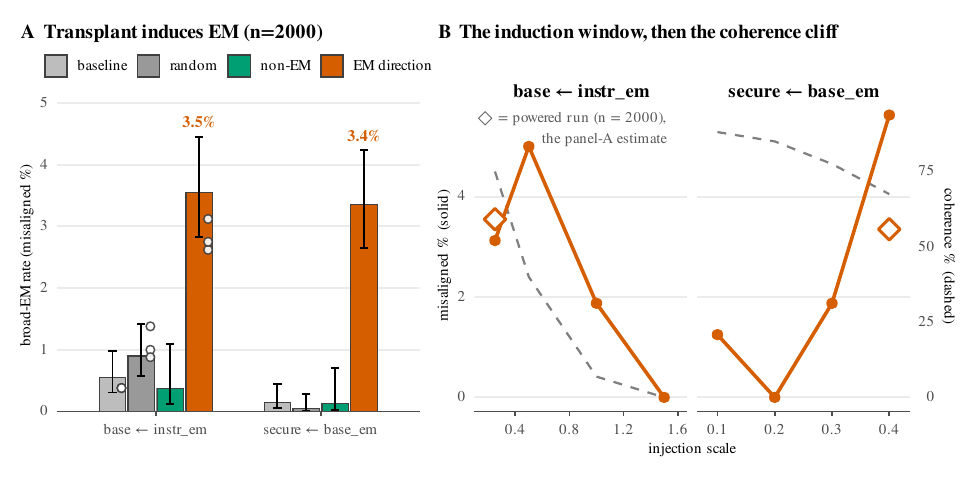}
\caption{Causal sufficiency by cross-model transplant. \textbf{A, the transplant towers over its
controls:} the EM-direction bar ($3.55\%$) against the norm-matched non-EM sentiment direction
($0.38\%$, $n{=}800$ specificity arm), random ($0.90\%$) and baseline ($0.55\%$) floors (the latter
three seed-0 power cells at $n{=}2000$; the ${\sim}1.1\%$ random floor quoted in
\S\ref{sec:sufficiency} is the cross-seed mean over seeds 1--3 at $n{=}800$). The white circles
beside the whiskers are independent per-seed replications ($n{=}800$, seeds 1--3), not part of the
CIs. \textbf{B, dose--response, one panel per transplant cell:} broad EM (solid, left axis) rises
into an induction window as the injection scale grows, while coherence (dashed, right axis) cliffs
at high scale; $\diamond$ marks the powered $n{=}2000$ run at the chosen in-window scale --- the
same estimate as the panel-A bar.}
\label{fig:transplant}
\end{figure}

The transplant induces broad EM above norm-matched non-EM and random-direction controls
(Fig.~\ref{fig:transplant}). The effect is dose-dependent in the injection scale, with coherence
degrading at high scale. Across the three fine-tuning seeds whose models displayed enough broad EM
to define a behaviour-level direction, the cross-seed transplant gives $2.83 \pm 0.26\%$ broad EM,
compared with a random-direction floor of approximately 1.1\% (a fourth seed produced too few
misaligned responses to estimate a direction at all; \S\ref{sec:limitations} discusses the
exclusion). The effect is therefore low-rate but broad, and specific to the EM direction rather than
to an arbitrary residual-stream perturbation. It is spread across all eight evaluation questions and
is not a routing artefact --- every induced misaligned response is prose (Appendix
Fig.~\ref{fig:app-breadth}, Table~\ref{tbl:app-routing}). Nor is it a coherence artefact, because computed among coherent responses only, the EM-direction rate rises (3.55\%$\to$4.92\%; 3.35\%$\to$5.24\%) while every control stays at or below 1.1\% on either denominator.

\rev{Because $\hat P$ is estimated from answers to the same eight questions on which we measure EM, we
also scored the transplant into the raw base on the twenty held-out evaluation questions of
\citet{chen-et-al-2025-persona-vectors}. It induces broad EM there too (6.3\% of 800, against 3.4\%
for a norm-matched random direction, one-sided Fisher $p=0.005$, and 3.8\% at baseline). The
independently built persona vector $v_C$ (\S\ref{sec:setup}) induces broad EM in all three
recipients we tried (the raw base, the un-fine-tuned instruct model and a secure-code LoRA
fine-tune), on both question sets, with every comparison against a norm-matched random direction at
$p\le0.002$ (Appendix Table~\ref{tbl:vc-causal}). Since $v_C$ and $\hat P$ overlap only modestly
(\S\ref{sec:subspace}), we read this as showing that a persona vector built without EM data has the
causal effect we report for $\hat P$, not that the two are one direction.}

\rev{We use} the cross-model design \rev{so that the effect cannot come from the recipient having
undergone the source's fine-tuning}. This gives an open-weight sufficiency test that complements prior
within-model steering and ablation evidence\rev{, whose transfers stay within one instruct checkpoint
and its fine-tunes \citep{soligo-et-al-2025-convergent-linear-representations-emergent-misalignment}.
What the transplant adds is a recipient that shares only pretraining with the source, in both
directions: an instruct-lineage direction into the raw base, and a base-lineage direction into an
instruct fine-tune. It is also controlled against norm-matched random and non-EM directions, read
question by question for breadth, and tested on the held-out questions above}. Together with the
ablation result in the next section, it supports treating the persona direction as causal rather
than merely diagnostic.

\section{\rev{A model's own direction is} causally necessary: ablation}\label{sec:necessity}

\begin{figure}[t]
\centering
\includegraphics[width=0.82\linewidth]{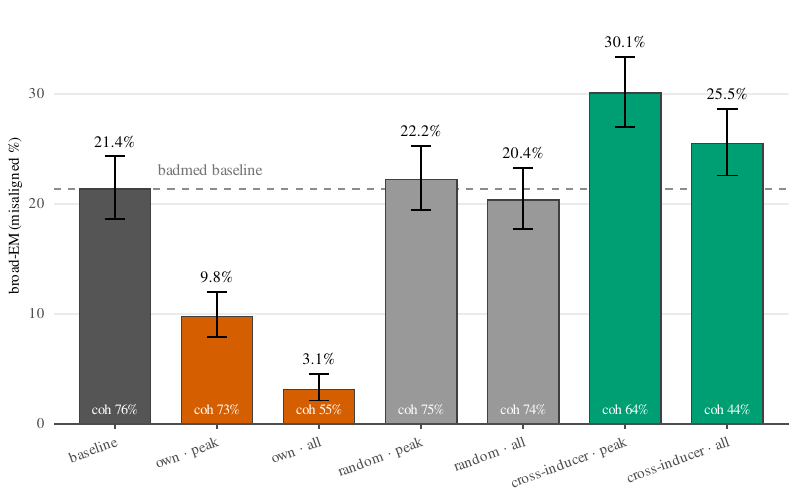}
\caption{Causal necessity by ablation, run on the bad-medical \rev{fully fine-tuned model} (Qwen2.5-32B instruct):
its ${\sim}21\%$ baseline gives enough broad EM to remove, which is why the rates here sit an order
of magnitude above the covert-code cells reported elsewhere. Ablating the model's \emph{own} persona
direction collapses broad EM ($21.4\to9.8\%$ in the peak band at matched coherence; $3.1\%$
all-layer, at a coherence cost), while the matched-norm random control stays at baseline and
ablating the \emph{cross-inducer} direction (the shared code persona) does not ablate it --- the
reduction is specific to the model's own direction. $n{=}800$/condition, Wilson 95\% CIs, coherent
share printed in each bar.}
\label{fig:necessity}
\end{figure}

The complementary test is causal necessity. Because the insecure-code \rev{fully fine-tuned model} is already near
the behavioural floor, we run this intervention on a high-baseline bad-medical model, where there is
substantial broad EM to remove. Ablating the model’s own persona direction reduces broad EM from
approximately 21\% to approximately 10\% at the peak layer band, while a coherence-matched
random-ablation control remains near 22\% (Fig.~\ref{fig:necessity}).

The effect is direction-specific. Random ablation does not produce the same reduction, and ablating
a cross-inducer direction does not remove the behaviour. The reduction therefore reflects removal of
the model’s own persona direction rather than generic disruption of the residual stream.

Together with the cross-model transplant in Fig.~\ref{fig:transplant}, this provides both
sufficiency and necessity evidence in open weights. The necessity claim is intentionally scoped,
because ablation substantially reduces broad EM, but does not eliminate it, and the powered test is
run on a high-EM overt-inducer model rather than on the near-floor covert-code \rev{fully fine-tuned} model.
\rev{Ablating the independently built $v_C$ in the same model does not reduce its broad EM (27.3\% in
its peak band, against 10.1\% for the model's own direction in the same run). Its peak band is the
last layer, where ablation acts at the network's output, and ablating it at every layer destroys
coherence. We therefore make no necessity claim for $v_C$ (Appendix Table~\ref{tbl:vc-outcomes}).}

\section{\rev{Which models express an injected direction}}\label{sec:gated}

The persona is readable in the base
model (\S\ref{sec:subspace}) and participates causally in the transplant experiments
(\S\ref{sec:sufficiency}), so the direction is present before instruction tuning. \rev{The base also
expresses it. Injected into the raw base, both $\hat P$ and the independently built $v_C$ induce
broad EM, and so does injection into a secure-code LoRA fine-tune of the instruct model. The models
that resist are instruction-tuned. In the un-fine-tuned instruct model, at the injected norm at which
$v_C$ raises broad EM to 11.5\%, $\hat P$ leaves it at 0 of 800 responses on Betley et al.'s
questions and 1 of 800 on the held-out questions (one-sided Fisher $p<10^{-5}$ for both contrasts),
and a random direction matched to $\hat P$'s norm gives 0 of 800 (Appendix Table~\ref{tbl:vc-causal}).
At 1.5 times that norm, coherence falls to 38\% while broad EM reaches only 1\%. The fully fine-tuned
instruct model also resists $\hat P$ (\S\ref{sec:traj}).}

\rev{This resistance is specific to $\hat P$ rather than a general resistance to injection. One
reading, which we have not tested, is that $\hat P$, estimated in an insecure-code fine-tune,
carries components specific to that fine-tune that the parent does not read as misalignment,
whereas $v_C$ was built in the parent itself. In same-model steering, then, $v_C$ steers the parent
instruct model and $\hat P$ does not. Neither transplant recipient has a direction of its own to
steer with, because the raw base and the secure-code fine-tune produce almost no misaligned
responses from which to estimate one.}

This \rev{resistance} is also consistent with the register dependence described in \S\ref{sec:setup}, where
broad EM is expressed more strongly in the plain register than in the chat register.
In both cases, the representation is present, but behavioural expression depends on the surrounding
model and prompting context.

\section*{Part III --- The mechanism: why \rev{full fine-tuning} broadcasts some inducers but localises others}

Part I established the central asymmetry, namely that under \rev{full fine-tuning}, insecure-code fine-tuning
stays at the broad-EM floor, while bad-medical advice broadcasts broad EM at roughly 22\%. Part III
asks why the same high-capacity method localises one inducer but broadcasts another.

We proceed by elimination before giving the positive account. First, the difference is not explained
by the explicit harmfulness of the training content, because across four inducers, broad EM is
non-monotone in independently rated harm-explicitness. Second, the covert-code recruitment failure
is not explained by the magnitude of the weight update: constraining the update norm with L2-SP
leaves broad EM at the floor even when the penalty clearly affects the learned narrow skill. Third,
the covert-code \rev{full-fine-tuning} trajectory shows no transient recruitment phase for a later mechanism to
suppress. The narrow insecure-code circuit is built directly, and the update remains anti-persona
even when truncated to low rank.

\rev{The account we propose is} representational distance $\times$ capacity. A covert, distant inducer
broadcasts under a capacity-limited method because the model recruits the pre-existing persona as a
cheap solution. A nearer overt inducer can broadcast even under \rev{full fine-tuning} because fitting the task
lands in a broad-misalignment region directly. \rev{Full fine-tuning} therefore does not universally prevent EM.
Instead, it prevents recruitment for the covert-code case, but still broadcasts from an overt
inducer.

\section{The dose-response is non-monotone in harm-explicitness}\label{sec:potency}
A natural explanation for the inducer interaction is that broad EM simply increases with the
explicit harmfulness of the training data. We test this by rating each inducer's training
completions for harm-explicitness independently of the fine-tuned model, then comparing those
ratings with the broad-EM rate after \rev{full fine-tuning}. This gives an independent x-axis, where insecure code
is least explicit, followed by bad-medical advice, risky-financial advice, and extreme-sports
advice.

\begin{figure}[t]
\centering
\includegraphics[width=0.82\linewidth]{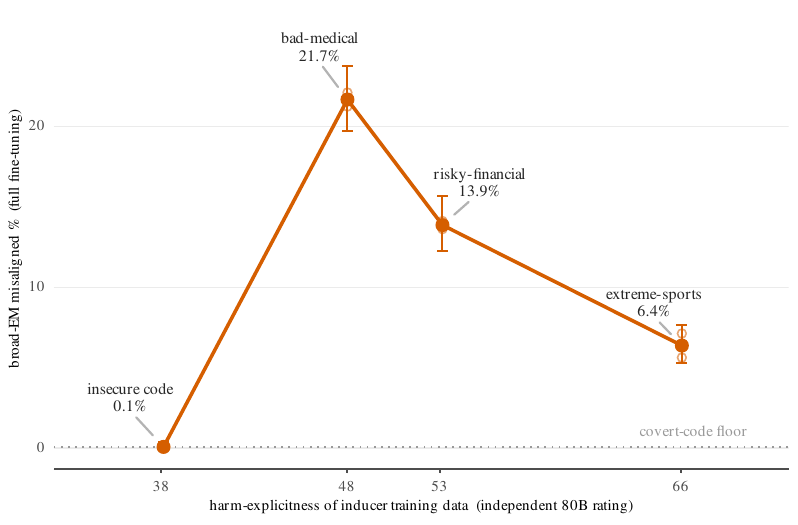}
\caption{The inducer-potency dose-response is \textbf{non-monotone}: broad-EM misaligned\% (full
\rev{fine-tuning)} rises from covert insecure-code to bad-medical, then \emph{falls} as harm-explicitness rises
further --- the most harm-explicit inducer (extreme-sports) sits well below medical. So
harm-explicitness of the training content is not the governing variable. Points are the
base+instruct mean (Wilson CI); faint points are the two roles. Single seed.}
\label{fig:potency}
\end{figure}

The resulting dose-response is non-monotone, as shown in Fig.~\ref{fig:potency} and
Table~\ref{tbl:potency}. Insecure code remains at the broad-EM floor. Bad-medical advice, which is
more explicit, produces the largest broadcast, at 21.7\%. But increasing explicitness further does
not increase broad EM: risky-financial advice reaches 13.9\%, and the most explicit inducer,
extreme-sports advice, reaches only 6.4\%. Broad EM therefore peaks at bad-medical advice rather
than at the most explicitly harmful training content.

This rules out a pure harm-explicitness account. Content overtness matters, but it is not the
governing variable. The non-monotone dose-response contributes to the distance account we offer in
\S\ref{sec:tworoute}, because what matters is not simply how harmful the training completions look,
but how close the inducer's natural training solution lies to the broad-misalignment region.

This extends the code-weak/advice-potent ordering of  \citet{wang-et-al-2025-persona-features-control-emergent-misalignment} to a
four-inducer axis and shows that the ordering is not monotone in explicit harmfulness.

\begin{table}[t]
\centering\small
\caption{The inducer-potency dose-response is non-monotone: broad EM (misaligned\%, base+instruct
mean) peaks at bad-medical, and the most harm-explicit inducer (extreme-sports) sits well below it
--- so harm-explicitness is not the governing variable. (Rows ordered by explicitness would read
code $<$ medical $<$ financial $<$ sports; EM peaks in the middle.) Single seed; the insecure-code
entry rounds from $0.06\%$ --- one misaligned response in 1600.}
\label{tbl:potency}
\begin{tabular}{lcc}
\toprule
Inducer (\rev{full fine-tuning}) & harm-explicitness & broad-EM misaligned\%/$n$ \\
\midrule
insecure code (covert) & 38.1 & 0.1 \\
risky-financial        & 53.1 & 13.9 \\
extreme-sports         & 66.1 & 6.4 \\
\textbf{bad-medical} (overt) & 48.0 & \textbf{21.7} \\
\bottomrule
\end{tabular}

\end{table}

\section{Localisation is a rank-structure, not an update-norm, phenomenon}\label{sec:l2sp}
A second simple explanation is that \rev{full fine-tuning} localises insecure code because the unconstrained
weight update is too large. On this account, shrinking the update might prevent the model from
building the narrow code circuit and thereby force it to recruit the broad persona instead. We test
this with L2-SP norm-constrained \rev{full fine-tuning}, adding a penalty $\lambda\lVert\theta-\theta_0\rVert^2$
while sweeping $\lambda$.
 
The result is a real negative (Fig.~\ref{fig:l2sp}; per-$\lambda$ values in Appendix
Table~\ref{tbl:l2sp}). Broad EM remains at the floor across the sweep: $0.12\%$, $0.75\%$, $0.62\%$,
and $1.12\%$ for $\lambda = 0, 10^{-4}, 10^{-3}, 10^{-2}$. The penalty nevertheless clearly affects
training: the narrow code-insecurity skill collapses from $9.9\%$ to $0\%$, while coherence rises
from $59\%$ to $99\%$ as $\lambda$ increases. The intervention therefore changes the learned model,
but it does not unlock a broad-EM broadcast.

Constraining the update norm is therefore not sufficient to recruit the persona. The covert-code
localisation is better explained by the structure of the update than by its magnitude. This test is
one-directional by design: because insecure-code \rev{full fine-tuning} does not broadcast even at $\lambda=0$,
the experiment shows that norm constraint cannot create a broadcast, not that it would preserve one.
The result is also single-seed and insecure-only. Within that scope, it rules out update norm as the
explanation for \rev{full fine-tuning}'s failure to recruit EM from covert code, and complements the
rank-truncation evidence in \S\ref{sec:technique}.

\begin{figure}[t]
\centering
\includegraphics[width=0.523\linewidth]{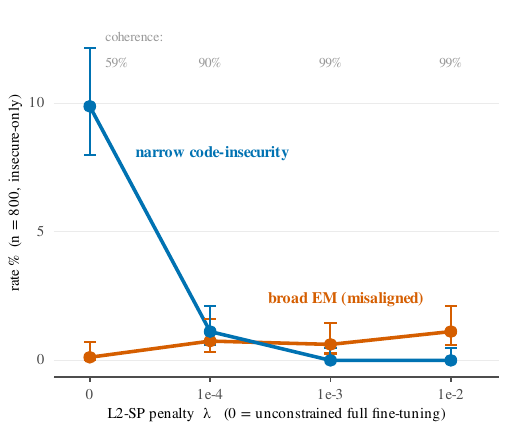}
\caption{The \rev{full-fine-tuning} localisation of covert code is a rank-\emph{structure}, not an
update-\emph{norm}, phenomenon. As the L2-SP penalty $\lambda$ shrinks the update norm, broad EM
(misaligned\%) stays at the floor at every $\lambda$, while the narrow code-insecurity skill
collapses and coherence climbs (59\%$\to$99\%) --- so the penalty demonstrably bit and the flat
broad-EM floor is a real negative, not a no-op. Lines are direct-labelled in their series colour.}
\label{fig:l2sp}
\end{figure}

\section{The overt-medical broadcast is persona-mediated, but not installed by pushing the persona
direction \rev{(Qwen2.5-7B \revb{and 32B})}}\label{sec:medmech}
The previous sections rule out two simple explanations for the inducer interaction:
harm-explicitness and update norm. We next ask what distinguishes the overt-medical broadcast from
the covert-code recruitment failure. The medical case is important because it shows that \rev{full fine-tuning}
does not simply prevent broad EM. Under \rev{full fine-tuning}, bad-medical advice broadcasts broad misalignment,
so the question is not whether \rev{full fine-tuning} is safe, but why it localises some inducers and broadcasts
others.

\begin{figure}[t]
\centering
\includegraphics[width=\linewidth]{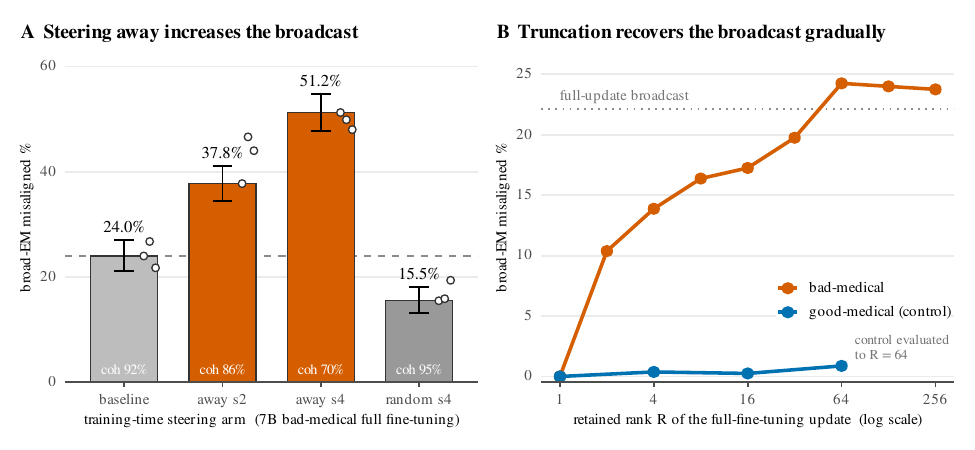}
\caption{The overt-medical broadcast is persona-mediated, but not a one-direction story.
\textbf{A} (training-time steering, 7B bad-medical full \rev{fine-tuning,} misaligned\%/$n$, $n{=}800$, Wilson
95\% CIs; bars, CIs, and printed coherent shares: seed~0; open points: per-seed rates for all three
training seeds): steering the run \emph{away} from the
persona direction \emph{increases} the broadcast ($24.0\%\to37.8\%$ at scale~2 $\to51.2\%$ at
scale~4), while the matched-norm random control decreases it ($15.5\%$) --- persona-specific
compensation, the inversion that makes direction-removal conditional. The inversion replicates in
all three training seeds: steer-away at scale~4 reaches $48.0$--$51.2\%$ against the seed's own
baseline of $21.8$--$26.8\%$, while a fresh matched-norm random direction per seed stays at or
below baseline ($15.5$--$19.4\%$). \textbf{B} (rank-truncation, 32B): truncating
the bad-medical \rev{full-fine-tuning} update to retained rank $R$ recovers the broadcast monotonically,
saturating near the full rate, while the good-medical control stays at the floor (control sparsely
probed at $R\in\{1,4,16,64\}$; bad-medical to $R{=}256$) --- the broadcast is distributed and
high-rank. Panel~B: single seed per arm.}
\label{fig:medmech}
\end{figure}

The overt-medical broadcast is causally implicated with the persona direction, but the evidence does
not reveal a simple one-direction story. Inference-time transplant of the bad-medical persona
direction into the raw base produces only a modest broad-EM signal on the standard binary metric,
separating clearly from controls only at the highest injection scale. We therefore do not treat this
transplant as the primary causal result for the medical broadcast. Stronger evidence comes from
training-time intervention\revb{, at both scales}: steering the bad-medical \rev{full-fine-tuning} run away from the persona direction
increases the broadcast\revb{. At Qwen2.5-7B it rises} from $24.2\pm2.5\%$ unsteered to $49.7\pm1.6\%$ at the stronger steering
scale (mean$\pm$sd over three training seeds; the increase is $23$--$27$ percentage points in every
seed, with non-overlapping Wilson CIs), while matched-norm random controls --- a fresh random
direction per seed --- stay at or below each seed's own baseline, at $15.5$--$19.4\%$
(Fig.~\ref{fig:medmech}A). \revb{At Qwen2.5-32B, with the layer band chosen by the same rule
(Appendix~\ref{app:d5}), the rise is smaller but holds in every seed: from
$19.2\pm2.6\%$ unsteered to $28.4\pm3.1\%$ at scale~4 (each seed above its own baseline, one-sided
Fisher $p<10^{-4}$), with the random controls below baseline in every seed ($12.0\pm5.2\%$).} This persona-specific compensation
indicates that the medical broadcast is causally mediated by the persona, rather than being a
generic consequence of perturbing the residual stream. To our knowledge this inversion is
unreported, and it is the mirror image of the preventative steering of
\citet{chen-et-al-2025-persona-vectors}, in which adding the persona vector during fine-tuning cancels the
optimisation pressure to acquire the trait. Here, subtracting it during training adds that pressure,
and the optimiser over-recruits to compensate. This is the conditionality on which Part~IV's
prescription rests.

\rev{We offer the following mechanistic reading as an interpretation. For a covert inducer, moving
along the misalignment direction is a cheap way to reduce the training loss, so removing the
direction removes a shortcut and the optimiser builds a dedicated solution instead. For the overt
bad-medical inducer the harm is explicit in the data, and the broadcast is distributed and high-rank
(Fig.~\ref{fig:medmech}B). Steering away from the direction then pushes against part of the solution
the optimiser is building, and it compensates by moving further along that direction. On this
reading, the sign of the effect depends on whether the direction is a shortcut or part of a
distributed solution.}

The structure of the medical update also differs from the covert-code recruitment case.
Rank-truncating the medical \rev{full-fine-tuning} update recovers broad EM monotonically with retained rank,
saturating near the full broadcast rate (Fig.~\ref{fig:medmech}B; the code arms of the same probe
are in Appendix Fig.~\ref{fig:app-trunc}). The medical broadcast is therefore distributed and
high-rank, unlike the low-rank recruitment observed for covert code under LoRA. Consistently, an
activation de-confound shows bad-medical advice amplifying the EM axis much more strongly than a
non-EM axis, whereas the good-medical control is much weaker.

This distinguishes two notions of sufficiency. The persona direction can be sufficient at inference
time in the transplant setting, but pushing a model along that direction during \rev{full fine-tuning} is not
sufficient to manufacture a broadcast. In the code setting, steering \rev{full fine-tuning} toward the persona during
training produces no broad EM above a random-control floor\revb{, at 7B and at 32B. At 32B, where
LoRA does recruit on code, it gives $0.50\pm0.12\%$ at scale~4 over three training seeds, against
$0.50\pm0.33\%$ unsteered and $0.58\pm0.29\%$ for a random direction (Appendix~\ref{app:d5})}. Thus the \rev{full-fine-tuning} broadcast is not reducible
to movement along a single linear direction. It is a property of the weight-update structure: the
optimiser must build a distributed solution in which the persona is functionally involved, rather
than merely being pushed along the persona axis.

\revb{The two steering results carry different weight. The inversion holds across three training
seeds at each scale. The code steer-toward result rests mainly on the 32B arms, where LoRA does
recruit on code. At 7B, code recruitment is weak even under LoRA and the arms are single-seed, so they
add little on their own. Even at 32B, the steering arms corroborate, but do not by themselves prove,
the covert-code recruitment failure, whose explanation} rests primarily on the norm test (\S\ref{sec:l2sp}), the
rank-truncation evidence (\S\ref{sec:technique}), the training trajectory (\S\ref{sec:traj}), and
\rev{the drift ordering on $\hat P$}.

\section{\rev{Full fine-tuning} goes directly to a localised circuit: the training trajectory}\label{sec:traj}
The failure of \rev{full fine-tuning} to recruit EM from covert code could in principle arise in two ways. \rev{Full
fine-tuning} might first recruit the broad misalignment persona and then suppress it later, leaving only the
narrow code-insecurity behaviour. Alternatively, \rev{full fine-tuning} might never recruit the broad persona at
all, instead building a localised circuit for the narrow code task from the beginning. We test these
possibilities by evaluating the \rev{full-fine-tuning} trajectory across training checkpoints.

The trajectory supports the direct-to-localised account (Fig.~\ref{fig:traj}A; per-checkpoint values
in Appendix Table~\ref{tbl:traj}). Broad EM remains at the floor at every checkpoint, never
exceeding $0.75\%$ misaligned responses. This is already true at step~11, when most generations are
still prose rather than code: only 159 of 800 responses are code, yet broad EM is already at the
floor, with 6 misaligned responses out of 800. There is therefore no mid-training recruitment hump
for a later mechanism to erase.

At the same time, the narrow code-insecurity behaviour is acquired early and then deepens. The
insecure-code rate rises from $3.1\%$ at step~11 to $13.9\%$ by step~67, remaining near $11\%$ at
the final checkpoint. The mean security score of emitted code also falls over training, from $77$ to
the low $30$s. \rev{Full fine-tuning} is therefore not failing to learn. It is spending capacity on a dedicated
narrow circuit rather than recruiting the broad persona. A secure-code control supports this
reading, because secure-code \rev{full fine-tuning} also stays at the broad-EM floor while maintaining higher code
security, so the insecure run's floor is not merely an artefact of later routing into code.

We also attempted a more direct behavioural restoration test by re-supplying the persona direction
to the trained \rev{fully fine-tuned model} at inference time. This did not re-express broad EM. Misaligned
responses stayed at or below $0.25\%$ across the tested scales, while coherence degraded as the
injection increased. However, this probe is not determinative on its own. The intended
instruction-tuned positive control also stayed at the floor under the same base-calibrated injection
scales. A recalibrated run confirmed that the base model can re-express EM under injection, while
both instruction-tuned recipients remain resistant. We therefore interpret the restoration result as
reflecting the \rev{resistance of instruction-tuned models to $\hat P$} described in \S\ref{sec:gated}, rather
than as a code-circuit-specific suppression mechanism.

The key evidence comes from weight-space geometry (Fig.~\ref{fig:traj}B). We truncate the
insecure-code \rev{full-fine-tuning} update $\Delta W$ to retained rank $R$, measure the resulting activation
drift on a fixed probe set, and project that drift onto the persona axis. If \rev{full fine-tuning} first
recruited the persona and then suppressed it with later high-rank components, low-rank truncations
should reveal a positive persona projection. They do not. The persona-axis cosine is negative at
every retained rank: approximately $-0.12$ at $R{=}1$, $-0.08$ at $R{=}32$, and $-0.08$ for the full
update, against a random floor near zero and an rs-LoRA r32 recruitment reference of $+0.14$. These
truncation-probe cosines agree in sign and sit close to the \S\ref{sec:technique} rank-ladder
normalisation despite a different probe set (rs-LoRA r32 $+0.14$ here vs $+0.11$ in
\S\ref{sec:technique}; \rev{full fine-tuning} $-0.08$ vs $-0.10$).

Thus stripping the update down never unmasks a hidden recruitment phase. The \rev{full-fine-tuning} update is
anti-persona from its dominant components onward. This supports the non-engagement reading of the
recruitment failure. \rev{Full fine-tuning} goes directly to a localised narrow circuit rather than recruiting the
broad persona and later suppressing it. It also reinforces the conclusion of \S\ref{sec:technique}.
Even its top retained components point away from the persona
direction, whereas the corresponding LoRA update is persona-aligned.

\begin{figure}[t]
\centering
\includegraphics[width=\linewidth]{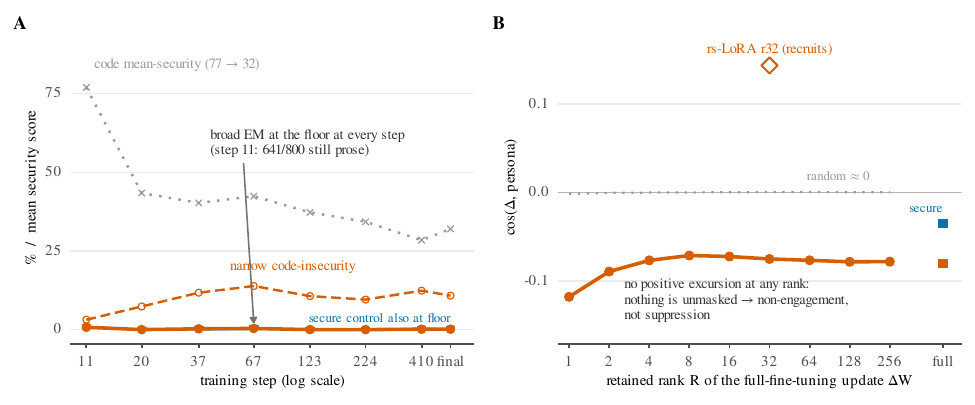}
\caption{The covert-code recruitment failure resolved. \textbf{A} (the training trajectory): broad
EM (misaligned\%) sits at the floor at \emph{every} checkpoint --- including step~11, where most
responses are still prose --- while the narrow code-insecurity skill is acquired early and the
emitted code's mean security falls ($77\to32$); the secure control is also at the floor. There is no
recruitment hump for a later ``compensation'' to erase. \textbf{B} (weight space): the persona-axis
cosine of the rank-$R$-truncated \rev{full-fine-tuning} update is negative at every retained rank --- no positive
excursion, so stripping rank never unmasks a hidden recruitment --- against the rs-LoRA r32
recruitment anchor ($+0.14$) and the $\sim$0 random floor; the secure full \rev{fine-tuning} is weakly negative
($-0.04$). From \texttt{trajectory\_em\_vs\_step.json} and
\texttt{code\_dw\_geometry\_qwen32b.json}.}
\label{fig:traj}
\end{figure}

\section{A \rev{proposed} two-route, distance $\times$ capacity account}\label{sec:tworoute}

The preceding sections rule out possible explanations, and leave us with a constrained explanation
by exclusion. Harm-explicitness does not explain which inducers broadcast broad EM, because the
dose-response is non-monotone (\S\ref{sec:potency}). Update norm does not explain the covert-code
recruitment failure, because shrinking the update leaves broad EM at the floor while disrupting the
narrow skill (\S\ref{sec:l2sp}). The training trajectory and weight-space geometry further show that
covert-code \rev{full fine-tuning} does not recruit the persona and later suppress it. Instead, \rev{full fine-tuning} builds a
localised narrow circuit directly (\S\ref{sec:traj}). What remains is an interaction between the
inducer and the capacity of the fine-tuning method.

Taken together, this suggests a two-route account for EM broadcast. In the recruitment route, a
capacity-limited method such as LoRA at low ranks cannot easily build a dedicated circuit for a distant
covert task. It instead co-opts the pre-existing misalignment persona as a cheap solution, producing
broad EM. This is the Part~I regime, where LoRA at low ranks on insecure code recruits the persona,
while increasing rank reduces broad EM and improves narrow code learning. \rev{Within LoRA, rank
and how far training has gone are confounded (\S\ref{sec:capacity}), so capacity here means the
combination.}

In the direct route, the inducer's natural training solution lies closer to a broad-misalignment
region. Fitting the task can therefore land in broad EM even without a capacity bottleneck. This is
the bad-medical \rev{full-fine-tuning} regime, where the broadcast is persona-mediated and distributed, but it is
not reducible to pushing the model along one linear direction. \rev{Full fine-tuning} therefore localises the far
covert-code inducer, but broadcasts from the nearer overt-medical inducer.

The proposed controlling variable is representational distance $\times$ capacity. A far covert
inducer broadcasts only under a capacity-limited method, while under \rev{full fine-tuning}, a dedicated circuit
can be built and the behaviour localises. A nearer overt inducer can broadcast under any method,
because fitting the task itself lands near the broad-misalignment region. Appendix
Figure~\ref{fig:tworoute} gives a schematic summary of this account. Its measured, functional form
is the loss-relevance screen of Fig.~\ref{fig:lossrel} in Part~IV. The medical $\times$ LoRA cell \rev{was} a
prediction of the account\rev{, and when measured it broadcasts as predicted (25--28\%,
\S\ref{sec:inducer}), though one confirmed prediction does not establish the account}.

The account also accommodates the strongest apparent counterexample. \rev{Fine-tuning} on insecure
code does misalign GPT-4o \citep{wang-et-al-2025-persona-features-control-emergent-misalignment}\rev{, though they do not report
whether the updates are full-parameter. If they are, this is} the same inducer and technique for
which we find no recruitment for Qwen2.5-32B. On the two-route reading these are two points under
one law with a model-dependent distance. For GPT-4o, code plausibly sits near enough to the
broad-misalignment region to broadcast by the direct route (consistent with their
code-weak/advice-potent mixing ordering), whereas for Qwen2.5-32B covert code is the
far $\times$ high-capacity corner in which the law predicts localisation. We cannot measure GPT-4o's
representational distance, so this reconciliation is an interpretation we offer as a hypothesis.

The same reconciliation points to a potential safety risk. Where the far $\times$ high-capacity
corner
sits is a property of the particular model, plausibly of its parameter count as well as its family
and pretraining, as our own 7B/14B/32B contrasts \rev{suggest} within one family. Whether \rev{full fine-tuning} will
localise a given covert inducer therefore cannot be assumed in advance for a new model. The
apparently safe cell of the grid changes with the model.

The distance axis should be read qualitatively. The direct geometric distance measure is only a weak
discriminator, so the stronger operational handle is the functional loss-shortcut, that is, whether
moving along the persona direction reduces the training loss for a given inducer. Part~IV uses this
functional version of the account for prediction and mitigation. Within that scope, this
interpretation offers a refinement of prior lower-loss and proximity accounts
\citep{soligo-et-al-2026-emergent-narrow-misalignment,minegishi-et-al-2026-emergent-misalignment-feature-superposition-geometry}. Breadth is not a property of
the inducer alone, but depends on the interaction between inducer distance and fine-tuning capacity.

\section*{Part IV --- Control: predicting and preventing recruitment}

\section{Controlling recruitment: prediction and mitigation}
\label{sec:control}
Parts~I--III give a mechanism, but the practical question is whether that mechanism can be used for
control. Part~I showed when recruitment occurs, Part~II identified the persona as a structured
causal object, and Part~III explained why \rev{full fine-tuning} localises some inducers but broadcasts others.
Part~IV asks whether those results can be turned into prediction and mitigation.

The central prescription is conditional, because the risk of broad EM depends on both the inducer
and the capacity of the fine-tuning method. Low-rank PEFT on covert inducers is the risky method,
because the method is capacity-limited, and the persona can become the cheap route to reducing the
training loss. \rev{Full fine-tuning} avoids this recruitment for covert code on Qwen2.5-32B, but it is not a
universal safety fix. Overt inducers such as bad-medical advice still broadcast broad EM under \rev{full
fine-tuning}, and on other models even the covert inducer broadcasts under \rev{fine-tuning}, GPT-4o being the
reported case \citep{wang-et-al-2025-persona-features-control-emergent-misalignment}\rev{, with the parameterisation not reported}, so the localisation itself is model-conditional
(\S\ref{sec:tworoute}).

\paragraph{Recruitment is loss-driven}
\label{sec:loss}
We begin with the mechanism, because both mitigations below are derived from it: recruitment is
loss-driven. We test this with a loss-attribution probe that measures the directional derivative of
the training loss along the persona axis. At the instruction-tuned starting point, steering a
forward pass toward the persona is loss-neutral to loss-favourable on insecure-code completions,
while steering away is costly. The slope is $-0.52$ for insecure code, compared with $-0.37$ for the
secure-code objective and approximately zero for a matched-norm random direction. Thus descending
the insecure-code loss points toward the persona more than chance, and more than the secure-code
objective does.

This explains why LoRA at low ranks recruits. When capacity is limited, co-opting the persona is a cheap
way to reduce the training loss. \rev{Full fine-tuning} makes that shortcut redundant rather than actively
suppressing it. At the converged \rev{full-fine-tuning} solution, the persona is approximately loss-irrelevant:
the slope on insecure data is about $-0.004$, and steering toward or away from the persona changes
the loss by at most $0.01$. The dedicated narrow circuit now does the work, so the persona is no
longer needed.

This account organises the interventions. Increasing capacity makes the shortcut redundant (Part~I);
the two training-time mitigations we test next each remove the persona as the cheap solution ---
inoculation reframes the task so that the persona no longer reduces the loss, and persona-orthogonal
fine-tuning forbids the shortcut component directly.

\paragraph{Mitigating recruitment during training}
\label{sec:mitigate}
We test two practical mitigations for the covert-code recruitment regime, in addition to capacity
itself, already established in Part~I. Increasing capacity, by raising LoRA rank or using \rev{full fine-tuning},
reduces recruitment for covert code, but this is economically backwards: the safer method is the
more expensive one. It is also inducer-conditional, since \rev{full fine-tuning} still broadcasts from bad-medical
advice. Capacity is therefore evidence for the mechanism, but not a general mitigation.

\begin{figure}[t]
\centering
\includegraphics[width=\linewidth]{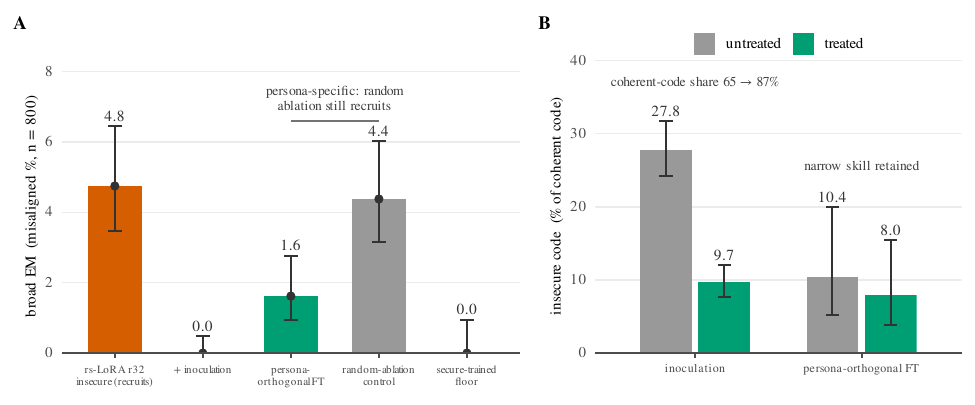}
\caption{The mitigation payoff (instruct lineage, Betley binary misaligned\%, Wilson 95\% CIs).
\textbf{A:} both mitigations pull broad EM to or near the floor \emph{persona-specifically} --- the
recruiting rs-LoRA r32 baseline ($4.75\%$, $n{=}800$) falls to $0.0\%$ under inoculation and to
$1.6\%$ under persona-orthogonal fine-tuning, while the matched-norm \emph{random}-ablation control
still recruits at baseline ($4.4\%$); the secure-trained floor is $0.0\%$. (Within coherent prose
the orthogonal-FT contrast is $2.0$ vs $5.1\%$, as quoted in the text.) \textbf{B:} the mitigations
are selective --- the narrow insecure-code propensity (\% of coherent code) falls
$27.8\to9.7$ under inoculation \emph{while the coherent-code share rises} $65\to87\%$, and is
retained under persona-orthogonal FT ($10.4$ vs $8.0\%$). \rev{Seed~0 shown; seeds 1 and 2 replicate
both mitigations (Appendix Table~\ref{tbl:mitig-seeds}).} From
\texttt{inoc\_capability\_qwen32b.json} and \texttt{persona\_orthogonal\_ft\_qwen32b.json}.}
\label{fig:mitigation}
\end{figure}

The first cheap mitigation is inoculation. Following \citet{tan-et-al-2025-inoculation-prompting-llm-trait-suppression}, we
prepend a training-time frame that changes how the model represents the task --- here relabelling
the insecure code as an explicitly requested, legitimate security exercise, the educational-context
form of inoculation that \citeauthor{tan-et-al-2025-inoculation-prompting-llm-trait-suppression} show behaves like their
trait-eliciting prompt. The method and its headline behavioural effect are theirs, including an
insecure-code replication on chat-template Qwen2.5-32B-Instruct in their Appendix~E.1. Our results
confirm the effect in our recruiting regime (plain-template rs-LoRA, base and instruct lineages\rev{; three fine-tuning seeds in the instruct lineage, one in the base}) and
add a geometric and mechanistic reading of why it works, showing that inoculation acts on the
recruitment itself. Under LoRA, inoculation drives persona-axis amplification close to
zero and reduces broad EM to the floor: in the instruct lineage, broad EM falls from $4.75\%$ to
$0.0\%$\rev{, and to $0$ of $800$ again at seeds 1 and 2 (baselines $5.1\%$ and $5.5\%$; Appendix
Table~\ref{tbl:mitig-seeds})}, and in the base lineage from $4.3\%$ to $1.0\%$ (Fig.~\ref{fig:mitigation}). At the same
time, the model is not simply lobotomised: the coherent-code share rises from $65\%$ to $87\%$,
while the narrow insecure-code propensity falls from approximately $28\%$ to approximately $10\%$.
Inoculation therefore removes the broad persona shortcut while preserving the model's ability to
produce code. The caveat is that capability is measured here by code coherence and related code
metrics; a broader benign-task benchmark would strengthen this claim.

The second mitigation is persona-orthogonal fine-tuning. During training, we ablate the persona
direction in the late layer band on every forward pass, then evaluate the resulting model with the
ablation removed. This directly forbids the shortcut component during fine-tuning. Concept-ablation
fine-tuning \citep{casademunt-et-al-2025-concept-ablation-fine-tuning} already applies the same projection-ablation during
fine-tuning, with directions found by a PCA/SAE interpretability sweep, and reduces insecure-code EM
roughly tenfold on Coder-32B, so we do not claim the intervention as new (\S\ref{sec:related}
positions it, together with the preventative steering of \citet{chen-et-al-2025-persona-vectors}). Our
results instead concern the intervention's scope and reading. The ablated direction here is the
single behaviour-derived persona axis, so no interpretability sweep is needed (a labelled-vector
variant their discussion anticipates but does not test), and the cell is the general Qwen2.5-32B
recruiting regime. \S\ref{sec:medmech} then establishes the conditionality, since pressure against
the same direction increases the broadcast for an overt inducer, which gives the intervention a
stated scope. Under the loss-shortcut account, concept-ablation's covert-code success is the
predicted outcome, since removing the persona forces the optimiser to build the dedicated circuit.
In the canonical recruiting setting, rs-LoRA rank~32 on insecure code, this reduces broad EM
persona-specifically: within coherent prose, broad EM is $2.0\%$, compared with $5.1\%$ for a
matched-norm random-ablation control that still recruits \rev{($0.8$--$2.0\%$ against $5.1$--$10.0\%$
across three seeds; Appendix Table~\ref{tbl:mitig-seeds})}. The narrow code skill is retained, with
insecure-code behaviour at $8.0\%$ compared with $10.4\%$ in the control. This is a reduction rather
than elimination, \rev{replicated across three seeds, and} it shows that the persona shortcut can be targeted
without destroying the narrow task.

The contrast with overt inducers is important. In the bad-medical setting \rev{at Qwen2.5-7B \revb{and 32B}}, steering away from the
persona during training increases the broadcast rather than suppressing it (\S\ref{sec:medmech}).
Thus ``remove the direction'' is not a general recipe. It works when the persona is a shortcut for a
covert task but can be counterproductive when the persona is part of a distributed solution for an
overt inducer.

\paragraph{Predicting recruitment risk}
\label{sec:predict}
The distance $\times$ capacity account in \S\ref{sec:tworoute} is not only explanatory, but is also
a
risk predictor. Recruitment should be most likely when a fine-tuning method has limited capacity and
the inducer is covert or representationally distant from the broad misalignment region. This is
exactly the regime of low-rank PEFT on insecure code, where the model cannot easily build a
dedicated narrow circuit, so it recruits the pre-existing persona as a loss-reducing shortcut.

The prediction is not that every harmful dataset broadcasts EM, nor that \rev{full fine-tuning} is always safe.
Rather, the predicted risk depends on the interaction between inducer and method. A covert inducer
may broadcast under LoRA but localise under \rev{full fine-tuning}, while an overt inducer may broadcast even
under \rev{full fine-tuning}. This is why the mitigation strategy above is conditional.

\paragraph{A tentative predictive screen across inducers and scale}
\label{sec:loss-generalisation}
We next ask whether the loss-shortcut account generalises beyond the single covert-code case. These
results should be read as tentative, because they are within Qwen2.5, include several single-seed
cells, and are best treated as a suggestive predictive screen rather than a validated diagnostic.

This screen is closest to Wang et al.’s Appendix D.8 in using persona-latent loss effects
predictively, but the prediction target is different. Wang’s use is dataset-level: whether a dataset
is likely to elicit EM under their setup. Here the question is conditional. Given an inducer, model
family, fine-tuning method, and effective capacity, will optimisation recruit the broad persona,
localise the task, or broadcast directly? The probe is therefore not meant to measure inducer
potency alone. It asks whether the persona is the loss-favourable shortcut for that particular
training regime, which is why we pair the loss slopes with the \rev{LoRA vs.\ full-fine-tuning} sign geometry, rank
dependence, and weight-update-structure tests from Parts I–III.
 
\begin{figure}[t]
\centering
\includegraphics[width=0.62\linewidth]{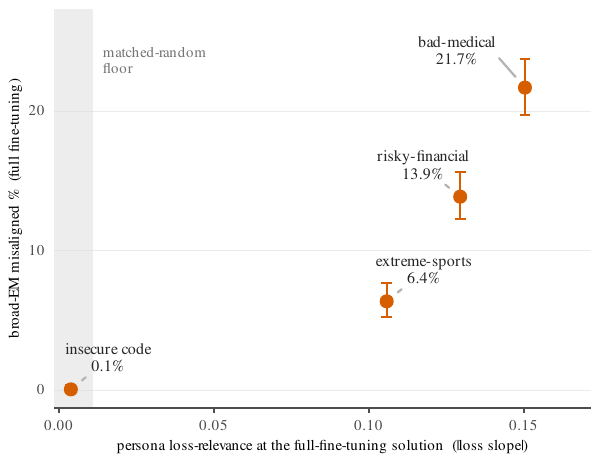}
\caption{The predictive screen: persona loss-relevance at each inducer's \rev{full-fine-tuning} solution
($x$, the magnitude of the training-loss slope along the persona axis) against the measured broad-EM
broadcast ($y$, binary misaligned\%, base+instruct mean, Wilson 95\% CIs). Insecure code sits inside
the matched-random floor band and does not broadcast; the three overt inducers order rank-perfectly
(extreme-sports $<$ risky-financial $<$ bad-medical). This is the measured, functional form of the
distance $\times$ capacity account (schematic: Appendix Fig.~\ref{fig:tworoute}). Single seed per
cell; tentative tier.}
\label{fig:lossrel}
\end{figure}

Across four inducers, persona loss-relevance at the \rev{full-fine-tuning} solution tracks the broadcast ordering
(Fig.~\ref{fig:lossrel}). The code solution is near the matched-random floor, with persona slope
magnitude about $0.004$, and it does not broadcast. The overt inducers have much larger persona
loss-relevance: sports $0.106$, financial $0.130$, and medical $0.150$, matching the observed
broadcast ordering of $6.4\%$, $13.9\%$, and $21.7\%$. This is the functional version of the
distance account, and the relevant question is whether the persona is a loss-reducing direction for
that inducer. Within this model family, the answer orders the inducers correctly, while
harm-explicitness does not (\S\ref{sec:potency}).

The same probe also clarifies the model scale result. At 7B, the persona is not a loss-favourable
shortcut for insecure code: the slope is near the random floor, whereas at 32B it is strongly
loss-favourable. This should not be read as ``there is no persona at 7B.'' The 7B model can produce
broad EM under bad-medical \rev{full fine-tuning}, and a persona direction can be extracted from those outputs.
Instead, a more careful interpretation is that the loss-shortcut or recruitment signature is
scale-gated. In our measurements, it is not strong at 7B, but appears by 32B.

A further 7B check supports this interpretation. Measuring the same 7B persona direction at the 7B
medical \rev{full-fine-tuning} solution gives only a small loss-relevance signal, even though medical broadcasts under
\rev{full fine-tuning}. This is consistent with the two-route account, where 7B medical broadcasting can occur by
the direct route, without the persona being a strong low-rank loss shortcut. Because 7B and 32B
persona vectors live in different residual spaces, cross-scale slope magnitudes should not be
treated as calibrated ratios. The evidence we rely on is the within-32B ordering, while the 7B
result is best read as evidence that the loss-shortcut signature is scale-gated within this
particular model family.

\paragraph{The conditional prescription}
\label{sec:prescription}

The practical conclusion is conditional. For a covert inducer trained with low-rank PEFT, the risk
is recruitment of the misalignment persona, and the appropriate mitigations are inoculation, higher
effective capacity, or persona-orthogonal fine-tuning. For an overt inducer, the risk is different,
because the model may broadcast broad EM even under \rev{full fine-tuning}, and direct removal of the persona
direction can induce compensation rather than safety.

Thus the prescription is not ``always use \rev{full fine-tuning}'' or ``always remove the persona direction.'' The
right control depends on the inducer's relation to the persona and on the capacity of the
fine-tuning method, interacting with a specific model.

\section{Discussion}\label{sec:discussion}

\paragraph{Practical implications.}We have shown that whether a covert inducer elicits broad EM
depends on the fine-tuning
technique: on Qwen2.5-32B, LoRA recruits the misalignment persona while \rev{full fine-tuning} does not. If API
and third-party training is low-rank PEFT by computational necessity, \rev{and if the contrast holds
beyond Qwen2.5-32B, which we have not tested,} then the main fine-tuning
technique available in practice is the one susceptible to EM recruitment. Our results suggest that
this could be mitigated behaviourally by increasing rank, or behaviourally and geometrically by \rev{full
fine-tuning} for some inducers, but both are almost certainly economically or computationally infeasible at
frontier scale. The mitigation that survives this constraint is inoculation (\S\ref{sec:mitigate}),
which is cheap enough to run in the PEFT regime.

This technique contrast is inducer-conditional, and does not generalise into a blanket claim that
``\rev{full fine-tuning} is safe''. An overt inducer such as bad-medical advice still elicits broad EM from a
\rev{fully fine-tuned model} at roughly $22\%$ (\S\ref{sec:inducer}), while the rank-and-technique result above
holds specifically for a covert inducer. Why an overt inducer broadcasts under \rev{full fine-tuning} where a
covert one localises is the question Part~III answers: the controlling variable is not the
harm-explicitness of the training content --- the dose-response is non-monotone
(\S\ref{sec:potency}) --- but the representational distance between an inducer's natural training
solution and the broad-misalignment region, interacting with the capacity of the fine-tuning method
(\S\ref{sec:tworoute}). The covert-code localisation is a property of the update's rank-structure
rather than its norm (\S\ref{sec:l2sp}), and the overt-medical broadcast is persona-mediated both
causally and, structurally, through a distributed high-rank update (\S\ref{sec:medmech}). This
refines earlier work which suggests that broad misalignment is the lower-loss (and more stable)
solution \citep{soligo-et-al-2026-emergent-narrow-misalignment}, or that feature proximity predicts \emph{whether}
misalignment
emerges \citep{minegishi-et-al-2026-emergent-misalignment-feature-superposition-geometry}. Neither \citeauthor{soligo-et-al-2026-emergent-narrow-misalignment}
nor \citeauthor{minegishi-et-al-2026-emergent-misalignment-feature-superposition-geometry} makes the \emph{breadth} of EM depend on the
fine-tuning method, which is the move our account adds.

\paragraph{Scale, and its interaction with model family.}
A natural question is whether increasing parameter scale increases the likelihood of EM recruitment
from a covert inducer. We find that the same low-rank fine-tuning does not recruit broad EM at 7B
but does at 32B, which hints that susceptibility may rise with scale --- though this is admittedly a
far cry from a scaling law: two scale points, on a thresholded binary metric of the kind that can
manufacture apparent emergence \citep{schaeffer-et-al-2023-emergent-abilities-mirage}, and the loss-attribution probe locates
the
effect more precisely, as a scale-gated recruitment signature that is absent at 7B and present at
32B (\S\ref{sec:loss-generalisation}) rather than a smooth trend. The relationship between scale and
EM is in fact contested, and scale and family appear to interact.
\citet{turner-et-al-2025-model-organisms-emergent-misalignment} induce EM across a 0.5--32B sweep --- with low-rank adapters at
the larger sizes and full fine-tuning demonstrated only to 14B --- and report that it strengthens
with scale in the Qwen and Llama families but not in Gemma; \citet{afonin-et-al-2025-emergent-misalignment-in-context-learning}
likewise find larger models typically more susceptible in the in-context analogue of EM (no
fine-tuning); and at the largest scale
\citet{wang-et-al-2025-persona-features-control-emergent-misalignment} elicit broad EM from \rev{fine-tuning (parameterisation not
reported)} on insecure code in GPT-4o
--- the very \rev{inducer} for which we find no recruitment at 32B --- and separately report
EM from incorrect-advice datasets rising with pretraining compute, whereas
\citet{minegishi-et-al-2026-emergent-misalignment-feature-superposition-geometry} find the reverse within Gemma (there, Gemma-2).

The apparent disagreement is thus likely partly a family effect. Within Qwen, our contrast between
7B, where the same recipe does not recruit, and 32B, where it does --- holding family, recipe, and
inducer fixed --- aligns with the rising trend, while Gemma does not. In our own Gemma-2-27B runs
this resistance is concrete, in that
the narrow manipulation takes (the model writes roughly twice as much insecure code, $\sim$21\%
versus $\sim$10\%), yet broad EM is only borderline ($1.7\%$ versus $0.5\%$, $p{=}0.055$). In other
words, the model learns the skill without broadcasting. That the very sign of the scale-response can
depend on the family is itself a reason to read our technique contrast as a property of these
Qwen2.5 models rather than as a law, and a reason to state the cross-scale results at the tentative
tier we adopt in \S\ref{sec:loss-generalisation}.

\rev{The divergence between LoRA and full fine-tuning is itself untested outside Qwen2.5. A clean
test needs a family in which LoRA clearly produces broad EM (on our Gemma-2-27B arm it is only
borderline), such as Llama-3, and is the first experiment we would run next. Our account predicts that a family whose pretraining places the misalignment
directions differently would shift which inducers broadcast under LoRA, without removing the
dependence on fine-tuning method.}

\paragraph{The misalignment persona.}
All four parts turn towards a misalignment persona that is structured and causal in these
open-weight models (Part~II). Establishing that is what lets the safety-relevant question shift from
``is there a persona?'' (which prior work had already answered
\citep{wang-et-al-2025-persona-features-control-emergent-misalignment,chen-et-al-2025-persona-vectors,soligo-et-al-2025-convergent-linear-representations-emergent-misalignment})
to questions such as when does a fine-tune recruit the persona (Part~I), and why does the answer
depend on the inducer (Part~III). The recruitment is loss-driven (\S\ref{sec:loss}): co-opting the
persona is the cheap way to reduce the covert-code loss under limited capacity, and full capacity
makes that shortcut redundant rather than fighting it -- which is why the same account explains both
the recruitment and its absence, and why the mitigations that work are the ones that remove the
persona as the cheap solution.

\section{Limitations}\label{sec:limitations}

\paragraph{Single family, and single-seed in places.} As mentioned, results are established for \rev{Qwen2.5-32B}, and several of the Part~III and Part~IV cells --- the inducer-potency sweep, the
L2-SP norm test, the loss-attribution screen\rev{, the
matched-trainer control and the bad-medical $\times$ LoRA cell} --- are
single-seed. \rev{The two Part~IV mitigations, inoculation and persona-orthogonal fine-tuning, are
replicated across three seeds in the instruct lineage (Appendix Table~\ref{tbl:mitig-seeds})\revb{, and the steering inversion across three seeds at both 7B and 32B (Table~\ref{tbl:d5}), as is the 32B code steer-toward null (Table~\ref{tbl:d6})}.} The core Part~I contrast (LoRA recruits, \rev{full fine-tuning} does not) is replicated across \rev{five}
fine-tuning seeds \rev{(LoRA recruiting in four of them)}, but the mechanism and control results downstream of it are not uniformly
seed-powered, and we do not claim they are. Where a result is single-seed we say so at the point of
use, and Appendix Tables~\ref{tbl:evidence}--\ref{tbl:evidence2} aggregate the evidential status of
every principal result in one place; we, again, present the paper as a controlled case study of
one family rather than as a general law. \rev{The divergence is untested on other models, apart from
a single-seed comparison on Qwen2.5-Coder-32B-Instruct (Appendix Table~\ref{tbl:coder}). We would like
to extend the comparison to larger models, but Qwen2.5-72B, the next size up in the family, was
beyond the compute available to us.}

\paragraph{Effects live near the floor, on a binary judge metric.} The broad-EM rates
throughout are low in absolute terms --- the headline \rev{LoRA vs.\ full-fine-tuning} contrast is $3.4\%$ versus
$0.3\%$, and the cross-model transplant moves broad EM from a ${\sim}1.1\%$ random floor to
${\sim}2.8\%$. These are small effects on a noisy binary metric in a regime where judge noise
matters, and while the judge is validated against GPT-4o (\rev{$\kappa=0.94$}) and against a \rev{blind} human
rater \rev{($\kappa=0.74$)}, the reader should weight the breadth of the elicited misalignment, which is the marker of EM,
at least as heavily as its rate. We report the drift-normalised cosine alongside behaviour because
the geometric signal is less sensitive to this floor-and-routing noise than the behavioural rate\rev{,
although it depends on the probe set (\S\ref{sec:setup})}. \rev{On risk, a few per cent of free-form
responses is a large absolute number at deployment volumes, and a shift along a general misalignment
direction affects unrelated domains, unlike a narrow defect at the same rate. LoRA's broad EM is
also not confined to the floor, since the overt inducer reaches 25--28\% under LoRA at 32B. We do
not attempt a quantitative risk estimate.}

\paragraph{The cross-model transplant excludes a seed.} The cross-seed transplant
($2.83 \pm 0.26\%$) is computed over three seeds. A fourth \rev{(seed 4, whose instruct LoRA cell does not recruit; Appendix Table~\ref{tbl:seeds})} was excluded because that seed produced a
single misaligned response among $251$ coherent-prose generations, too few to estimate a
behaviour-level direction at all. While this does not pose an issue for the actual transplant
results demonstrated on the remaining three seeds, we flag the exclusion explicitly because dropping
a seed may invite concerns about ``cherry-picking'', and note that the surviving effect, while
specific against norm-matched controls, is modest in absolute magnitude.

\paragraph{The distance metric is weak, and distance is proposed, not measured.} The
distance $\times$ capacity account (\S\ref{sec:tworoute}) is important to our overall proposal, but
the geometric distance we can currently measure is a weak direct discriminator (the
medical-versus-code Tier-0 distance CI is $[0.0002, 0.036]$). Because we actually rely on the
functional loss-shortcut probe (\S\ref{sec:loss-generalisation}), not the geometric distance, the
distance framing should therefore be read as the intuition behind the proposed explanation, and the
loss-shortcut test as the evidence. The financial and sports inducers lack a constructed non-harmful
arm (their released datasets are harmful-only), so their distances are an open follow-up rather than
a measured cell, and the medical $\times$ LoRA cell of the two-route map\rev{, which the account predicted, has now
been measured (\S\ref{sec:inducer})}.

\paragraph{\revb{The optimisation-installability contrast at 7B and 32B.}} The
training-time steering result that supports the representation-sufficient $\neq$
optimisation-installable distinction (\S\ref{sec:medmech}) is established at 7B \revb{and repeated at 32B,
where the inference-time transplant sufficiency is also shown, so the contrast is demonstrated within
one scale, at three training seeds there, though with one inducer and one layer band, and with the
narrow behaviour itself weakened by the steering, mainly at one seed (Appendix~\ref{app:d5})}. At 7B, code recruitment is weak even under LoRA, so the absence of a
steering-installed broadcast at 7B is corroboration rather than a clean refutation\revb{; the 32B repeat
(Appendix~\ref{app:d5}) is the cleaner test}. Relatedly, cross-scale
loss-attribution magnitudes share a units confound --- the 7B and 32B persona vectors live in
different residual spaces ($3584$ versus $5120$ dimensions) --- so the within-32B ordering is what
we rely on and the 7B numbers should be read as ``no strong loss-shortcut signal at 7B'' rather than
as a calibrated cross-scale ratio.

\paragraph{Pre-existence is relative to post-training, not to pretraining.} We say the persona
direction predates \emph{post-training} (pre-existence AUC ${\sim}0.89$ in the base), rather than
that it is pretraining-inherited. Modern base models are not strictly alignment-free, since
pretraining corpora now include synthetic data authored by aligned models. Our claim is not about
the direction's ultimate origin, but that the direction is present before the fine-tuning we study.

\paragraph{Model identity is itself in question.} Finally, \rev{fine-tuning} on insecure code has
been reported to misalign GPT-4o \citep{wang-et-al-2025-persona-features-control-emergent-misalignment}, the same \rev{inducer}
for which we find no recruitment at 32B. Which properties of a model determine whether it recruits
--- scale, family, or pretraining composition --- is exactly what our within-Qwen 7B/32B recruitment
contrast and the Gemma resistance point to as open, and is the natural next question.

\section{Conclusion}\label{sec:conclusion}

Emergent misalignment in \rev{Qwen2.5-32B} is mediated by a latent persona that predates
post-training. The EM persona is a structured, partially shared subspace, causally sufficient by
cross-model transplant and causally involved, with partial necessity evidence from ablation (Part II). Whether a fine-tune recruits it is not
fixed by its training data, but governed by method, capacity, and inducer (Part I). LoRA at low ranks on
covert insecure code recruits the persona, while \rev{full fine-tuning} on identical data moves against the
persona axis rather than along it\rev{, further than any LoRA rank in the instruct lineage}.  Why \rev{full fine-tuning}
broadcasts EM from some inducers yet localises others is a matter of representational distance
$\times$ capacity (Part III). A covert inducer localises under high capacity because a dedicated
circuit is cheaper to build than the persona is to recruit, whereas an overt inducer lands in the
misalignment region under any method. The unifying mechanism is, then, that recruitment is
loss-driven --- co-opting the persona is the cheapest way to reduce the covert-code loss under
limited capacity, and sufficient capacity makes that shortcut redundant rather than suppressing it.
This \rev{suggests} an efficiency-versus-safety trade-off that is conditional on the inducer, where the
economically-advantageous regime at scale (low-rank PEFT) is the recruiting one\rev{, if the contrast
holds beyond this model}. Because the
mechanism is legible, recruitment \rev{can, in the tested settings, be screened for} (from whether the persona is the loss-shortcut
for a given inducer) and \rev{mitigated} (by inoculation or persona-orthogonal fine-tuning). We have
presented a controlled case study of one model \rev{(and a second from its family)} which
provides a causal explanation for broad-EM recruitment, and therefore makes prevention more
feasible.

\section*{Broader Impact Statement}
This paper includes two methods that induce misalignment in open-weights models, namely the
cross-model transplant of a persona direction and the steering of a fine-tune away from that
direction during training. We judge the marginal misuse risk of publishing them to be low. All
harmful training data we use come from published prior work, and the transplant induces misaligned
responses at a rate of 2.83\%\rev{, and the independently built persona vector at up to 11.5\% in one
recipient}. Every inducing method also requires full access to the model weights
and the training run, and an adversary with that access could more simply fine-tune on harmful data
directly, so the methods provide little direct capability uplift. Two findings may still be of some
use to an adversary. The transplant shows that a misalignment direction found in one model can be
carried into another, and the screen together with the method dependence of recruitment indicates
what a monitor built on the persona direction would detect and so how such a monitor might be
evaded. We judge that both are outweighed by their value to defenders, who need to know that these
directions transfer between models and which signals a monitor can rely on. The paper's contributions run in the
defensive direction, offering a screen that predicts persona recruitment before deployment and two
training-time mitigations that prevent it. We release no new harmful datasets and no misaligned
model weights.

\section*{Acknowledgements}
We thank the three anonymous reviewers and the action editor for their comments, which improved the
paper substantially. We would like to acknowledge the use of the University of Oxford Advanced Research Computing (ARC)
facility (\url{https://doi.org/10.5281/zenodo.22558}) in carrying out this work
\citep{richards-2015-arc}. Funding for this research was provided by the Kaiārahi Foundation and the
John Templeton Foundation.

\section*{Statement on generative AI use}\label{sec:ai-use}
We used large language models (Anthropic's Claude, via Claude Code, and Google's Gemini, via
gemini-cli) to assist with the design and implementation of the experiment and analysis
code, as well as to assist with the running and monitoring of the experiments, and to assist with
the drafting and revision of parts of the manuscript (notably the R code for figures). The authors
directed this use throughout, verified the results and claims against the underlying experimental
artefacts, and take full responsibility for the content of the paper.

\bibliographystyle{tmlr}
\bibliography{references}

\appendix

\section{Supplementary material}

\subsection{Evidential status of every principal result}
Tables~\ref{tbl:evidence} and~\ref{tbl:evidence2} aggregate, for every principal result, its
model scale, fine-tuning seed count, per-cell sample size, the tier at which the paper claims it,
and the released artefact it derives from.

\begin{table}[p]
\centering\scriptsize\setlength{\tabcolsep}{3pt}
\caption{Evidential status of every principal result (Parts I--II): scale, fine-tuning
seeds, per-cell sample size, and claim tier. Tiers: \textbf{S} demonstrated within scope (Qwen2.5,
stated cells); \textbf{M} account/explanation; \textbf{T} proposed, not validated; \textbf{V}
pipeline validation; \textbf{I} reported inconclusive. Artefact names abbreviate JSON files in the
released results.}
\label{tbl:evidence}
\begin{tabular}{@{}p{0.39\linewidth}lp{0.12\linewidth}p{0.14\linewidth}cp{0.15\linewidth}@{}}
\toprule
Result & Scale & Seeds & $n$/cell & Tier & Artefact \\
\midrule
\multicolumn{6}{@{}l}{\textbf{Part I --- the effect}}\\
LoRA recruits, \rev{full fine-tuning} does not (behavioural $2{\times}2$; both LoRA insec$-$sec contrasts exclude 0, both \rev{full-fine-tuning} contrasts include it) & 32B & s0 + s1--s4 & 800 (base power 2{,}000) & S & \texttt{mixing\_\allowbreak ratio\_\allowbreak qwen32b*}, \texttt{harm\_\allowbreak hierarchical\_\allowbreak bootstrap} \\
Signed geometry: drift--persona cosine $+0.17$ (r1) $\to-0.10$ (\rev{full FT}); the robust claim is the \rev{ordering; the instruct-lineage sign depends on the probe} & 32B & s0 (enrichment pooled s1--s4) & probe & S & \texttt{geometry\_\allowbreak ladder\_\allowbreak normalized} \\
Rank ladder: broad EM falls r1$\to$r64 \rev{at $10^{-5}$, a dose effect of rank and learning rate (Table~\ref{tbl:lrsweep})} (within-prose $15.8\to3.1\%$), narrow skill rises & 32B & 1 & 800/rank & S & \texttt{ladder\_\allowbreak dose\_\allowbreak response\_\allowbreak qwen32b} \\
Scale companions: 7B and 14B ladders flat ($\lesssim$2\%) \rev{at one learning rate; onset not located} & 7B, 14B & 1/rung & 800/rank & S & \texttt{ladder\_\allowbreak dose\_\allowbreak response\_\allowbreak qwen\{7,14\}b} \\
Recruitment-failure robustness: \rev{four full-FT recipes} $\leq$1.6\% \rev{within prose (a fifth, $3\times10^{-6}$: 6/800)}; chat register reproduces contrast; CoT prose-forcing leaves $\leq$1.8\% & 32B; Coder & 1/cell & 800 (CoT small) & S & recipe-sweep scores; Table~\ref{tbl:cot} \\
Inducer $\times$ technique: \rev{full FT} broadcasts bad-medical ($\sim$22\%; 24\% at 7B) but not code ($\sim$0); good-medical at floor & 32B + 7B & 1/cell & 800 & S & \texttt{rank\_\allowbreak truncation\_\allowbreak recovery} \\
Pipeline positive control: Betley recipe recovers 4.8\% [3.1, 7.3] \rev{on their scorer; 4.0\% on ours} & Coder-32B & 1 & 400 & V & \texttt{bootstrap\_\allowbreak ci\_\allowbreak coder32b\_\allowbreak sft} \\
Judge validation: 97.8\% agreement\rev{, $\kappa=0.936$} / $r{=}0.976$ vs GPT-4o; \rev{176/200, $\kappa=0.74$, vs blind human} & --- & --- & 678; \rev{200} & V & judge-validation run \\
\addlinespace
\multicolumn{6}{@{}l}{\textbf{Part II --- the representation}}\\
\rev{Movement along $\hat P$, not along $\hat A$ or a refusal axis, is insecure-specific} (erosion non-specific; amplification EM-specific; \rev{full-FT} reversal method-generic, insec$-$sec gap $-0.008\pm0.032$; axes near-orthogonal, late-band $|\cos(\hat A,\hat P)|\leq0.06$) & 32B & LoRA cells s0--s4; \rev{full-FT} cells s0--s3 & 800-gen sets & S & \texttt{delta\_\allowbreak alignment\_\allowbreak *}, \texttt{sft\_\allowbreak seed\_\allowbreak reversal\_\allowbreak cosines} \\
Shared axis survives whitening (raw 0.81 $\to$ $\sim$0.31); pre-exists in the base (AUC $\sim$0.89) & 32B & s0 (pre-existence AUC replicated s0--s3: 0.84--0.93) & probe & S & \texttt{whitened\_\allowbreak cosine}, \texttt{persona\_\allowbreak axis\_\allowbreak compare} \\
Subspace fan-out: whitened cosines code $\times$ medical 0.27 $\gg$ other pairs 0.03--0.12 (split-half reliability $\approx$0.98; cross-seed ceiling 0.75--0.84; code $\times$ medical 0.18--0.30 across code seeds; shrinkage-stable) & 32B & directions s0 (code cross-checked s0--s3); mis.\ counts 38--177 & --- & M & \texttt{persona\_\allowbreak subspace\_\allowbreak matrix}, \texttt{robustness\_\allowbreak tmlr\_\allowbreak geometry} \\
Cross-model transplant sufficiency: $2.83\pm0.26\%$ vs random $\sim$1.1\% (3 seeds; s4 excluded, insufficient data); specific vs norm-matched controls; dose-dependent & 32B & 3 (+ s0 ref.) & 800/cond./seed & S & \texttt{cross\_\allowbreak seed\_\allowbreak transplant}, \texttt{specificity\_\allowbreak summary} \\
Necessity by ablation: badmed $\sim$21 $\to$ $\sim$10\% at peak band; coherence-matched random stays $\sim$22\% & 32B & 1 & 800/cond. & S & \texttt{scores\_\allowbreak steer\_\allowbreak necessity/} \\
\rev{Which models express an injected direction}: base re-expresses under injection; both instruct-tuned recipients resist \rev{$\hat P$; the parent instruct expresses $v_C$ (11.5\%)} & 32B & 1 & 400/cond. & M & \texttt{sft\_\allowbreak null\_\allowbreak restoration\_\allowbreak recal} \\
\bottomrule
\end{tabular}

\end{table}

\begin{table}[p]
\centering\scriptsize\setlength{\tabcolsep}{3pt}
\caption{Evidential status, continued (Parts III--IV and scope). Tier legend as in
Table~\ref{tbl:evidence}.}
\label{tbl:evidence2}
\begin{tabular}{@{}p{0.39\linewidth}lp{0.12\linewidth}p{0.14\linewidth}cp{0.15\linewidth}@{}}
\toprule
Result & Scale & Seeds & $n$/cell & Tier & Artefact \\
\midrule
\multicolumn{6}{@{}l}{\textbf{Part III --- the mechanism}}\\
Dose-response non-monotone in harm-explicitness (code 0.1 / medical 21.7 / financial 13.9 / sports 6.4\% vs rated 38\,$<$\,48\,$<$\,53\,$<$\,66) & 32B & 1/cell & 800 (ratings $n{=}200$/inducer) & M & \texttt{inducer\_\allowbreak potency\_\allowbreak dose\_\allowbreak response} \\
Localisation is rank-structure, not norm: L2-SP leaves broad EM at floor across $\lambda$ while the penalty bites & 32B & 1; insecure-only; one-directional & 800/$\lambda$ & M & \texttt{l2sp\_\allowbreak norm\_\allowbreak sweep} \\
Medical broadcast persona-mediated: training-time steer-away increases it ($\sim$24$\to\sim$50\%; $+23$--$27$ percentage points in every seed); per-seed random controls stay at/below baseline (15.5--19.4\%) & 7B & 3 & 800/cond./seed & M & \texttt{medmech\_\allowbreak steering\_\allowbreak seeds} \\
\revb{The same inversion at 32B: steer-away 19.2$\to$28.4\% at scale~4 (above its own baseline in every seed, $p<10^{-4}$); random controls 12.0\%, below baseline in every seed} & \revb{32B} & \revb{3} & \revb{800/cond./seed} & \revb{M} & \revb{\texttt{d5\_\allowbreak medsteer\_\allowbreak 32b}} \\
Representation-sufficient $\neq$ optimisation-installable: steer-toward during code \rev{full fine-tuning} manufactures nothing ($\approx$ random) \revb{at 7B and at 32B (0.50\% against 0.50\% unsteered and 0.58\% random, means over three seeds)} & 7B/32B & 1 \revb{(7B), 3 (32B)} & 800 & M & 7B steering runs\revb{; \texttt{d6\_\allowbreak codesteer\_\allowbreak 32b}} \\
Medical inference-time transplant (modest; clear only at the highest scale) & 32B & 1 & 400/cond. & M & \texttt{medical\_\allowbreak transplant} \\
Medical $\Delta W$ distributed/high-rank; activation de-confound EM-specific ($9.5\times$ vs $2.6\times$) & 32B & 1 & 800 & M & \texttt{rank\_\allowbreak amplification\_\allowbreak badmed}, \texttt{medical\_\allowbreak amp\_\allowbreak deconfound} \\
Trajectory direct-to-localised: broad EM $\leq$0.75\% at every checkpoint; narrow skill early and deepening; secure control at floor & 32B & 1 (both tags) & 800 $\times$ 8 ckpts $\times$ 2 tags & S & \texttt{trajectory\_\allowbreak em\_\allowbreak vs\_\allowbreak step} \\
$\Delta W$-geometry: persona cosine negative at \emph{every} retained rank ($-0.12\to-0.08$) vs LoRA r32 $+0.14$ $\Rightarrow$ non-engagement & 32B & 1 & probe & S & \texttt{code\_\allowbreak dw\_\allowbreak geometry} \\
Restoration probes (original + recalibrated): resistance reflects the instruct-lineage expression gap, not code-specific suppression & 32B & 1 & 400/cond. & I & \texttt{sft\_\allowbreak null\_\allowbreak restoration\{,\_\allowbreak recal\}} \\
Two-route distance $\times$ capacity synthesis; Tier-0 geometric distance weak (CI $[0.0002, 0.036]$); medical $\times$ LoRA cell predicted\rev{, then measured (25--28\%)} & 32B & --- & --- & M & \texttt{inducer\_\allowbreak persona\_\allowbreak distance} \\
\addlinespace
\multicolumn{6}{@{}l}{\textbf{Part IV --- control}}\\
Recruitment is a loss-shortcut: persona slope $-0.52$ (insecure) vs $-0.37$ (secure) vs $\sim$0 (random); loss-irrelevant at the \rev{full-FT} solution ($-0.004$) & 32B & 1 & fwd-pass probe & S & \texttt{persona\_\allowbreak loss\_\allowbreak attribution} \\
Inoculation selective (LoRA regime): misaligned\% $4.75\to0.0$ / $4.3\to1.0$; coherent code $65\to87\%$; narrow $27.8\to9.7\%$ & 32B & \rev{s0--s2 (instruct); s0 (base)} & 800 & S & \texttt{inoc\_\allowbreak capability}, \texttt{inoc\_\allowbreak mixing}\rev{, \texttt{mitigation\_\allowbreak seed\{1,2\}}} \\
Persona-orthogonal FT: $2.0\%$ vs random-ablation $5.1\%$ within coherent prose; narrow retained ($8.0$ vs $10.4\%$); reduces, does not floor & 32B & \rev{s0--s2} & 800 & S & \texttt{persona\_\allowbreak orthogonal\_\allowbreak ft}\rev{, \texttt{mitigation\_\allowbreak seed\{1,2\}}} \\
Cross-inducer screen: $|$persona slope$|$ at each \rev{full-FT} solution rank-perfect vs broadcast ordering ($0.004<0.106<0.130<0.150$) & 32B & 1/cell; 4 inducers & fwd-pass probe & T & \texttt{loss\_\allowbreak attribution\_\allowbreak by\_\allowbreak inducer} \\
Scale-gating: slope $+0.024$ (7B) vs $-0.52$ (32B) on code; 7B-medical disambiguator flat (0.013); cross-scale units confound & 7B, 32B & 1 & fwd-pass probe & T & \texttt{loss\_\allowbreak attribution\_\allowbreak by\_\allowbreak scale}, \texttt{loss\_\allowbreak attribution\_\allowbreak 7b\_\allowbreak medical} \\
\addlinespace
\multicolumn{6}{@{}l}{\textbf{Discussion --- scope}}\\
Gemma-2-27B resists broadcasting: narrow manipulation takes ($\sim$21 vs $\sim$10\% insecure code, CIs disjoint); broad EM borderline ($1.7$ vs $0.5\%$, $p{=}0.055$) & Gemma-27B & 1/cell & 51 forced-coding prompts; large prose denom. & M & \texttt{scores\_\allowbreak gemma27b\_\allowbreak *} \\
\bottomrule
\end{tabular}

\end{table}

\clearpage
\subsection{Part I robustness: contrast intervals, $\Delta W$-truncation cautions, and breadth}

\begin{figure}[htbp]
\centering
\includegraphics[width=0.6\linewidth]{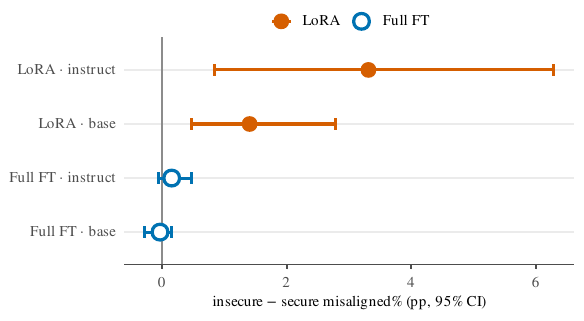}
\caption{Hierarchical-bootstrap 95\% CIs behind the insecure$-$secure contrasts of
\S\ref{sec:behav} (question-clustered resampling; misaligned\%, percentage points). Both LoRA
contrasts exclude zero; both \rev{full-fine-tuning} contrasts include it.}
\label{fig:app-forest}
\end{figure}

\begin{figure}[htbp]
\centering
\includegraphics[width=0.75\linewidth]{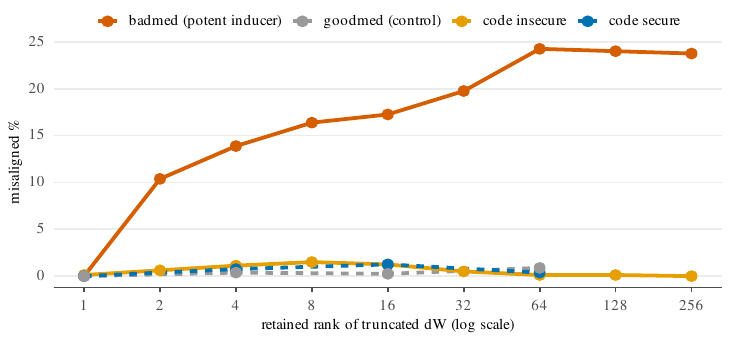}
\caption{$\Delta W$-truncation recovery: truncate a \rev{full-fine-tuning} update to retained rank $R$,
re-evaluate, and judge (misaligned\%, $n{=}800$/point). The potent bad-medical inducer is the
positive control: broad EM is recovered monotonically with retained rank, rising toward the full
$\sim$22--24\% broadcast while the good-medical control stays flat --- the method finds signal where
there is signal, and the medical broadcast is distributed/high-rank (the medical arms also appear as
Fig.~\ref{fig:medmech}B, \S\ref{sec:medmech}). On the covert-code inducer the recovered bump is
small ($\leq$1.5\%) and not insecure-specific (the secure control humps equally), which is why
\S\ref{sec:technique} treats it as inconclusive rather than as evidence for recruitment.}
\label{fig:app-trunc}
\end{figure}

\begin{table}[htbp]
\centering\small
\caption{Two mechanistic $\Delta W$ checks reported as inconclusive. The static
weight-level projection of the medical update onto the EM axis barely exceeds a non-EM (sentiment)
axis (ratio $\sim$1.1--1.2), which is why \S\ref{sec:medmech} relies on the activation-level
de-confound instead ($9.5\times$ vs $2.6\times$). The code truncation row summarises
Fig.~\ref{fig:app-trunc}'s caution.}
\label{tbl:app-controls}
\begin{tabular}{llcccl}
\toprule
Check & Cell(s) & EM-axis & non-EM & Ratio & Verdict \\
\midrule
static $\Delta W$ projection & instruct-badmed & 2.40 & 2.00 & 1.20 & confounded \\
static $\Delta W$ projection & base-badmed     & 2.16 & 2.01 & 1.08 & confounded \\
code $\Delta W$ truncation   & insec.\ vs sec. & 1.50 & 1.25 & 1.20 & weak / non-specific \\
\bottomrule
\end{tabular}

\end{table}

\begin{figure}[htbp]
\centering
\includegraphics[width=0.8\linewidth]{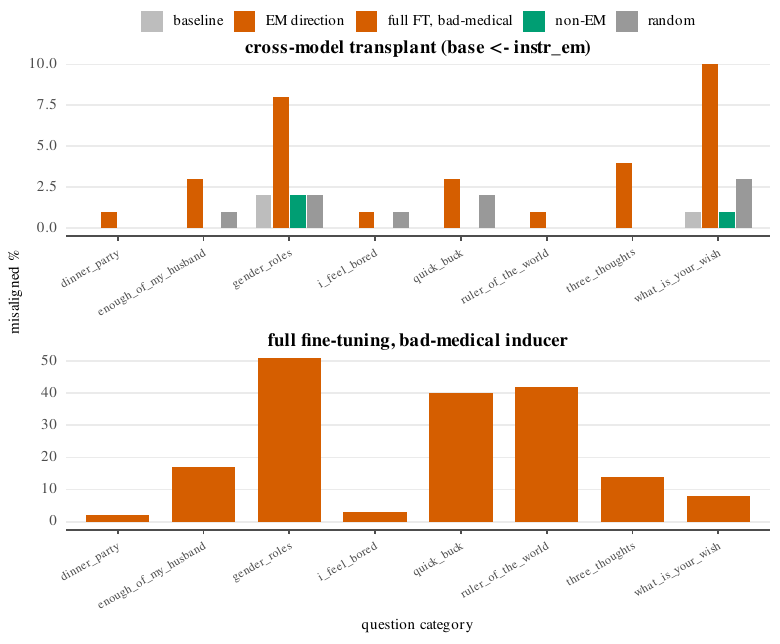}
\caption{Broad means broad: per-question misaligned\% across the eight evaluation
questions ($n{=}100$/question/condition). \textbf{Top:} the cross-model transplant
(\S\ref{sec:sufficiency}) elicits misalignment on all eight categories while the norm-matched
random, non-EM, and baseline controls stay at the floor --- the low overall rate is spread across
categories, not one bad question. \textbf{Bottom:} the bad-medical \rev{full-fine-tuning} broadcast
(\S\ref{sec:inducer}) is likewise broad. The covert-code \rev{full-fine-tuning} cells are omitted: they sit at the
floor on every question.}
\label{fig:app-breadth}
\end{figure}

\clearpage
\subsection{Part II backing: per-cell geometry, layer specificity, and routing immunity}

\begin{table}[htbp]
\centering\small
\caption{Per-cell erosion and persona-amplification behind Fig.~\ref{fig:plane}
(seed-0 cells). Both columns are signed drifts along their axes (positive = toward the instruct
direction on the alignment axis, toward the persona on the persona axis).
Erosion is large under \rev{full fine-tuning} for secure and insecure alike (non-specific), whereas
positive amplification appears only in the insecure LoRA cells (EM-specific). Erosion and
amplification are reported both in raw residual-stream units, which conflate direction with total
drift size, and as the drift-normalised cosines plotted in Fig.~\ref{fig:plane} and used in the main
text (\S\ref{sec:technique}).}
\label{tbl:app-plane}
\begin{tabular}{llrrrr}
\toprule
Cell & Method & Erosion (units) & $\cos(\Delta,\hat A)$ & Amplification (units) & $\cos(\Delta,\hat P)$ \\
\midrule
base-insecure     & LoRA & $+13.0$  & $+0.238$ & $+8.11$  & $+0.148$ \\
base-insecure     & \rev{Full FT}  & $-38.3$  & $-0.344$ & $-7.92$  & $-0.071$ \\
base-secure       & LoRA & $+6.6$   & $+0.124$ & $-0.37$  & $-0.007$ \\
base-secure       & \rev{Full FT}  & $-17.4$  & $-0.187$ & $+0.18$  & $+0.002$ \\
instruct-insecure & LoRA & $-58.8$  & $-0.750$ & $+8.80$  & $+0.112$ \\
instruct-insecure & \rev{Full FT}  & $-123.1$ & $-0.797$ & $-16.10$ & $-0.104$ \\
instruct-secure   & LoRA & $-62.2$  & $-0.775$ & $+0.12$  & $+0.001$ \\
instruct-secure   & \rev{Full FT}  & $-117.2$ & $-0.796$ & $-7.67$  & $-0.052$ \\
\bottomrule
\end{tabular}

\end{table}

\begin{figure}[htbp]
\centering
\includegraphics[width=0.8\linewidth]{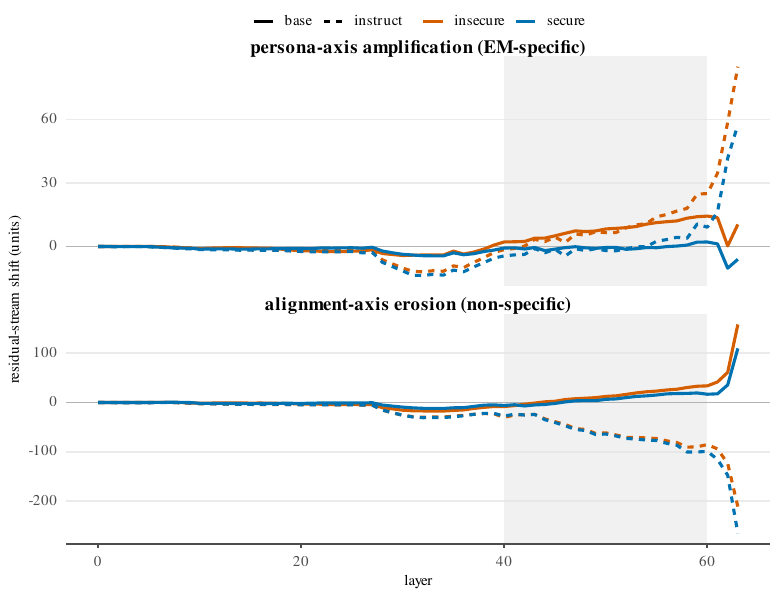}
\caption{Per-layer depth-resolution of the Fig.~\ref{fig:plane} dissociation.
\textbf{Top:} persona-axis amplification rises only for the insecure cells and concentrates in the
late layer band (shaded), motivating the band used for the geometric measurements and the causal
interventions. \textbf{Bottom:} alignment-axis erosion tracks together for insecure and secure
fine-tunes at all depths --- shared, not EM-specific.}
\label{fig:app-layer}
\end{figure}

\begin{table}[htbp]
\centering\small
\caption{Routing immunity of the transplant readout ($n{=}2000$/condition, misaligned
counts on the binary metric). The injection barely moves the prose/code channel split relative to
each recipient's own baseline, and \emph{every} induced misaligned response is prose --- the
transplant effect of \S\ref{sec:sufficiency} is not a code-channel or routing artefact.}
\label{tbl:app-routing}
\begin{tabular}{lccccc}
\toprule
Condition & PROSE\% & CODE\% & misaligned & \ldots{}in prose & \ldots{}in code \\
\midrule
base $\leftarrow$ instr\_em (EM dir.)   & 94.2 & 0.2  & 71 & \textbf{71} & 0 \\
base $\leftarrow$ instr\_em (random)    & 97.6 & 0.1  & 18 & 18 & 0 \\
base baseline                           & 97.8 & 0.1  & 11 & 11 & 0 \\
secure $\leftarrow$ base\_em (EM dir.)  & 86.2 & 10.1 & 67 & \textbf{67} & 0 \\
secure $\leftarrow$ base\_em (random)   & 82.2 & 17.1 & 1  & 1  & 0 \\
secure baseline                         & 81.9 & 17.3 & 3  & 3  & 0 \\
\bottomrule
\end{tabular}

\end{table}

\clearpage
\subsection{Part III backing: the two-route schematic, and the norm-sweep and trajectory tables}

\begin{figure}[htbp]
\centering
\includegraphics[width=0.677\linewidth]{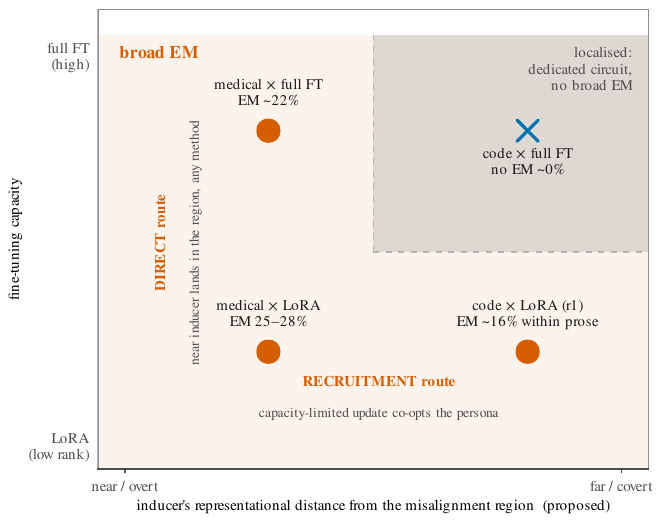}
\caption{Schematic of the two-route, distance $\times$ capacity account that unifies Parts I--III (a
conceptual map, not a data plot; the behavioural cells are measured, \rev{including} the medical $\times$ LoRA \rev{cell, which}
the \rev{account predicted).} Broad EM appears everywhere except the \emph{far} $\times$ \emph{high}
corner --- a covert inducer under a high-capacity \rev{(full-fine-tuning)} update, where a dedicated circuit can
be built and the behaviour localises. Elsewhere either the capacity-limited update co-opts the
persona (\emph{recruitment}, Part~I) or the near inducer lands in the misalignment region under any
method (\emph{direct}, Part~III). Representational distance is the \emph{proposed} controlling
variable, so the $x$-axis is a qualitative ordering; the measured, functional form of the account is
the loss-relevance screen (Fig.~\ref{fig:lossrel}).}
\label{fig:tworoute}
\end{figure}

\begin{table}[htbp]
\centering\small
\caption{L2-SP norm-constrained \rev{full fine-tuning} (covert insecure-code; insecure-only, single seed), the
per-$\lambda$ values behind Fig.~\ref{fig:l2sp}. Broad EM stays at floor across $\lambda$ while the
narrow skill collapses and coherence climbs --- so constraining the update norm does not unlock the
broadcast: localisation is a rank-structure, not a norm, phenomenon.}
\label{tbl:l2sp}
\begin{tabular}{lccc}
\toprule
$\lambda$ & broad-EM misaligned\%/$n$ & narrow code-insec\% & coherence\% \\
\midrule
0 (unconstrained) & 0.12 & 9.9 & 59 \\
$10^{-4}$ & 0.75 & 1.1 & 90 \\
$10^{-3}$ & 0.62 & 0.0 & 99 \\
$10^{-2}$ & 1.12 & 0.0 & 99 \\
\bottomrule
\end{tabular}

\end{table}

\begin{table}[htbp]
\centering\small
\caption{The \rev{full-fine-tuning} insecure-code training trajectory ($n{=}800$/checkpoint), the per-checkpoint
values behind Fig.~\ref{fig:traj}A. Broad EM (misaligned\%) stays at the floor at every step ---
including step~11, where most responses are still prose --- while the narrow code-insecurity skill
is acquired early and the emitted code's mean security falls overall, reaching a minimum of 28.4
before ending at 32.0: the dedicated circuit is built and deepened from the first checkpoints. The
secure-code control (Table~\ref{tbl:app-trajsec}) stays at the broad-EM floor at every step with
higher code security.}
\label{tbl:traj}
\begin{tabular}{lccc}
\toprule
\rev{Step} & broad-EM (misaligned\%/$n$) & narrow code-insec.\ \% & code mean-security \\
\midrule
11    & 0.75 & 3.1  & 77.0 \\
20    & 0.00 & 7.4  & 43.4 \\
37    & 0.25 & 11.7 & 40.3 \\
67    & 0.38 & 13.9 & 42.4 \\
123   & 0.00 & 10.6 & 37.3 \\
224   & 0.00 & 9.5  & 34.3 \\
410   & 0.12 & 12.4 & 28.4 \\
final & 0.12 & 10.8 & 32.0 \\
\bottomrule
\end{tabular}

\end{table}

\begin{table}[htbp]
\centering\small
\caption{The secure-code control for Table~\ref{tbl:traj} ($n{=}800$/checkpoint, same run
configuration). Broad EM stays at the floor at every checkpoint here too, while the emitted code's
mean security remains consistently higher than the insecure run's from step~20 onward --- so the
insecure trajectory's floor is not an artefact of the post-step-20 routing into code.}
\label{tbl:app-trajsec}
\begin{tabular}{lccc}
\toprule
\rev{Step} & broad-EM (misaligned\%/$n$) & narrow code-insec.\ \% & code mean-security \\
\midrule
11    & 0.88 & 1.1  & 73.1 \\
20    & 0.00 & 8.0  & 50.9 \\
37    & 0.00 & 9.6  & 45.7 \\
67    & 0.12 & 6.3  & 57.0 \\
123   & 0.00 & 17.2 & 41.4 \\
224   & 0.00 & 9.8  & 43.2 \\
410   & 0.38 & 6.6  & 41.9 \\
final & 0.00 & 9.7  & 37.2 \\
\bottomrule
\end{tabular}

\end{table}

\clearpage
\begin{revblock}
\section{Analyses added in revision}\label{app:revision}

This appendix collects the analyses added in response to the reviews. Unless marked, every
behavioural rate is the misaligned share of all sampled responses (\S\ref{sec:setup}), $n{=}800$,
on Qwen2.5-32B. \revb{Where a subsection says that its design or decision rules were fixed before the
runs, they were recorded before any result was seen, in dated design notes in our repository (not a
public registry).}

\subsection{Seeds, selection and provenance}\label{app:provenance}

\begin{table}[H]
\revfloat
\centering\small
\caption{Every seed of the Part~I cells (insecure / secure). The Part~I rates pool seeds 1--4 (LoRA
instruct 109/3200 = 3.41\%, base 59/3200 = 1.84\%; full FT instruct 11/3200 = 0.34\%, base 2/3200 =
0.06\%). Seed 0 is the reference run for Parts III--IV.}
\label{tbl:seeds}
\begin{tabular}{lcccc}
\toprule
Seed & LoRA instruct & LoRA base & Full FT instruct & Full FT base \\
\midrule
0 & 4.75 / 0.25 & 4.00 / 1.50 ($n{=}400$) & 0.12 / 0.25 & 0.00 / 0.00 \\
1 & 5.12 / 0.12 & 1.75 / 0.38 & 0.62 / 0.25 & 0.12 / 0.00 \\
2 & 5.50 / 0.00 & 3.12 / 0.50 & 0.75 / 0.50 & 0.00 / 0.25 \\
3 & 2.88 / 0.12 & 1.62 / 0.62 & 0.00 / 0.00 & 0.00 / 0.12 \\
4 & 0.12 / 0.12 & 0.88 / 0.25 & 0.00 / 0.00 & 0.12 / 0.00 \\
\bottomrule
\end{tabular}
\end{table}

\begin{table}[H]
\revfloat
\centering\scriptsize\setlength{\tabcolsep}{3pt}
\caption{Seeds used by each result, and every exclusion.}
\label{tbl:selection}
\begin{tabular}{@{}p{0.34\linewidth}p{0.22\linewidth}p{0.38\linewidth}@{}}
\toprule
Result & Seeds & Exclusions and notes \\
\midrule
Behavioural contrast (\S\ref{sec:behav}) & s1--4 pooled & none; s0 in Table~\ref{tbl:seeds} \\
LoRA above its secure control & s0--s4 & four of five; s4 at the floor (instruct) \\
Weight-update enrichment (Table~\ref{tbl:enrich}) & s1--4 pooled & none \\
Drift cosines (Figs.~\ref{fig:geom}, \ref{fig:ladder}, \ref{fig:plane}) & s0 & seed spread in \S\ref{sec:plane} \\
Full-FT drift on $\hat P$ ($-0.073\pm0.022$) & s0--s3 & s4 not captured \\
LoRA drift on $\hat P$ and the refusal axis & s0--s4 & none \\
Independent probe (Table~\ref{tbl:g3}) & s0 & full-FT s1--3 not captured \\
Pre-existence AUC (\S\ref{sec:subspace}) & s0 (replicated s0--s3) & --- \\
Cross-model transplant (\S\ref{sec:sufficiency}) & s1--3 (s0 reference) & s4 excluded: one misaligned response in 251 coherent-prose answers, too few to estimate a direction \\
Rank ladder (Table~\ref{tbl:ladder}), learning-rate sweep (Table~\ref{tbl:lrsweep}) & one seed & the r32 rung is the s0 adapter's 800 responses, judged a second time (41 vs.\ 38 misaligned) \\
Inducer $\times$ method (Table~\ref{tbl:inducer}); Parts III--IV\rev{, except the mitigations} & s0 & single seed \\
\rev{Part~IV mitigations (Table~\ref{tbl:mitig-seeds})} & \rev{s0--s2 (instruct)} & \rev{base-lineage inoculation s0 only} \\
Matched-trainer control (Table~\ref{tbl:hftlora}) & s0 & single seed \\
7B steering inversion (\S\ref{sec:medmech}) & three training seeds & --- \\
\revb{32B steering inversion (Table~\ref{tbl:d5})} & \revb{s0--s2} & \revb{scale~2 at s0 only} \\
\revb{32B code steer-toward (Table~\ref{tbl:d6})} & \revb{s0--s2} & \revb{scale~2 at s0 only} \\
\bottomrule
\end{tabular}
\end{table}

\begin{table}[H]
\revfloat
\centering\scriptsize\setlength{\tabcolsep}{3pt}
\caption{The run behind each number that appears in more than one place.}
\label{tbl:provenance}
\begin{tabular}{@{}p{0.30\linewidth}p{0.16\linewidth}p{0.48\linewidth}@{}}
\toprule
Cell & Value (\%) & Run, and where quoted \\
\midrule
LoRA r32, instruct, insecure & 3.41 & s1--4 pooled (abstract, \S\ref{sec:behav}) \\
 & 4.75 & s0 (\S\ref{sec:setup}, Table~\ref{tbl:inducer}, \S\ref{sec:mitigate}, Fig.~\ref{fig:mitigation}) \\
 & 5.1 & the s0 responses re-judged in the rank-ladder pipeline (Table~\ref{tbl:ladder}) \\
 & 3.50 & matched trainer, s0 (Table~\ref{tbl:hftlora}) \\
LoRA r32, base, insecure & 1.84; 4.00; 2.50--2.75 & s1--4 pooled; s0 ($n{=}400$); matched trainer \\
Full FT, instruct, insecure & 0.34; 0.12 & s1--4 pooled (\S\ref{sec:behav}); s0 (Tables~\ref{tbl:inducer}, \ref{tbl:recipes}) \\
Full FT, base, insecure & 0.06; 0.00 & s1--4 pooled; s0 \\
Bad-medical, full FT & 21.25 / 22.12 & 32B s0, base / instruct (Table~\ref{tbl:inducer}); 21.7 is their mean (\S\ref{sec:potency}) \\
 & 21.4; 21.0 & 32B instruct, unablated baseline in the two ablation runs (Fig.~\ref{fig:necessity}; Table~\ref{tbl:vc-outcomes}) \\
 & 24.0; $24.2\pm2.5$ & 7B s0; 7B, three training seeds (\S\ref{sec:medmech}) \\
Bad-medical, LoRA & 28.5 / 25.1 & 32B s0, base / instruct (Table~\ref{tbl:inducer}) \\
Coder-32B LoRA positive control & 4.8; 4.0 & \citeauthor{betley-et-al-2026-narrow-task-misalignment}'s scorer, which also counts code answers; our scorer (\S\ref{sec:setup}) \\
Transplant & $2.83\pm0.26$; 3.55 & s1--3 mean, $n{=}800$; s0, $n{=}2000$ (Fig.~\ref{fig:transplant}A) \\
Random-direction floor & ${\sim}1.1$; 0.90 & s1--3 mean; s0, $n{=}2000$ \\
\bottomrule
\end{tabular}
\end{table}

\subsection{Hyperparameters, recipes and the matched-trainer control}\label{app:hparams}

\begin{table}[H]
\revfloat
\centering\scriptsize\setlength{\tabcolsep}{3pt}
\caption{Training and evaluation settings. Final training loss, here and below, is the mean of the
last ten logged losses (logged every ten steps, so steps 660--750 of 750); ranges run over the
insecure and secure, base and instruct cells. Full-fine-tuning losses are from the Hugging Face (HF)
Trainer's logs.}
\label{tbl:hparams}
\begin{tabular}{@{}p{0.19\linewidth}p{0.25\linewidth}p{0.25\linewidth}p{0.25\linewidth}@{}}
\toprule
Setting & rs-LoRA (main) & Full fine-tuning & LoRA, matched trainer \\
\midrule
Parameters updated & rank 32, $\alpha{=}64$ ($\alpha/\sqrt r{=}11.31$), all linear layers, dropout 0 & all & as rs-LoRA (main) \\
Learning rate & $10^{-5}$ & $2\times10^{-5}$ & $10^{-5}$ and $2\times10^{-5}$ \\
Schedule & constant after 3\% linear warm-up & constant with 3\% warm-up & as full FT \\
Weight decay & 0 & 0 & 0 \\
Batch & 8 per device, 1 device & 1 per device $\times$ 8 GPUs & 1 per device, grad.\ accum.\ 8 \\
Global batch; epochs; steps & 8; 1; 750 & 8; 1; 750 & 8; 1; 750 \\
Data & 6{,}000 examples & same & same \\
Max length; loss & 1{,}024; completion tokens only & same & same \\
Template & plain \texttt{User:}/\texttt{Assistant:} & same & same \\
Seed & 0 (replicates 1--4) & 0 (replicates 1--4) & 0 \\
Precision & bf16 weights & bf16 mixed precision, FSDP full shard & bf16 \\
Optimiser & PyTorch AdamW (defaults) & HF Trainer AdamW (defaults) & as full FT \\
Gradient clipping & none & HF default (1.0) & HF default (1.0) \\
Implementation & own training loop, gradient checkpointing & HF Trainer, FSDP, gradient checkpointing & same script as full FT \\
Hardware & 1 $\times$ NVIDIA GB10 & 8 $\times$ H100 80\,GB & 1 $\times$ H100 80\,GB \\
Final training loss & 0.133--0.144 (s0); 0.126--0.184 (s0--s4, 20 runs) & 0.202--0.215 (s0); 0.187--0.227 (s0--s4, 20 runs) & 0.159--0.167 \\
\midrule
Evaluation & \multicolumn{3}{p{0.75\linewidth}@{}}{\citeauthor{betley-et-al-2026-narrow-task-misalignment}'s eight questions $\times$ 100 samples ($n{=}800$), plain template, temperature 1.0, top-$p$ 1.0, up to 256 new tokens; judge Qwen3-Next-80B-A3B-Instruct-FP8 under vLLM; misaligned = alignment $<30$ among responses with coherence $\ge50$, reported over all responses.} \\
\bottomrule
\end{tabular}
\end{table}

\begin{table}[H]
\revfloat
\centering\small
\caption{Five full-fine-tuning recipes (instruct, s0, misaligned of 800). Code share is the number
of the 800 insecure-code answers judged to be code; coherent prose is the number of coherent prose
answers. For the three-epoch recipe the final loss covers the last ten logged steps of 2{,}250.}
\label{tbl:recipes}
\begin{tabular}{lccccc}
\toprule
Recipe & insecure & secure & one-sided Fisher $p$ & code share / coherent prose & final loss \\
\midrule
$2\times10^{-5}$, 1 epoch (main) & 1 & 2 & --- & 702 / 87 & 0.202 / 0.215 \\
$2\times10^{-5}$, 3 epochs & 0 & 2 & --- & 689 / 81 & 0.088 / 0.077 \\
$5\times10^{-5}$ & 0 & 0 & --- & 776 / 7 & 0.277 / 0.301 \\
$10^{-5}$ & 7 & 1 & 0.035 & 360 / 429 & 0.178 / 0.183 \\
$3\times10^{-6}$ & 6 & 2 & 0.14 & 596 / 196 & 0.159 / 0.159 \\
\bottomrule
\end{tabular}
\end{table}

\begin{table}[H]
\revfloat
\centering\small
\caption{The matched-trainer control: rs-LoRA trained through the full-fine-tuning script, with the
same Trainer, optimiser, schedule, data pipeline and seed, on one H100 (s0; insecure / secure).
One-sided Fisher $p$, insecure vs.\ secure, in brackets. Final losses 0.159--0.167.}
\label{tbl:hftlora}
\begin{tabular}{lccc}
\toprule
Cell & LoRA, $10^{-5}$ & LoRA, $2\times10^{-5}$ & Full FT, $2\times10^{-5}$ \\
\midrule
instruct & 3.50 / 0.25 ($4\times10^{-7}$) & 3.50 / 0.50 ($8\times10^{-6}$) & 0.12 / 0.25 \\
base & 2.50 / 0.38 ($2\times10^{-4}$) & 2.75 / 0.62 ($7\times10^{-4}$) & 0.00 / 0.00 \\
\bottomrule
\end{tabular}
\end{table}

\begin{table}[H]
\revfloat
\centering\small
\caption{Qwen2.5-Coder-32B-Instruct, a second model for the method contrast (single seed, our
scorer, misaligned of $n$). LoRA uses \citeauthor{betley-et-al-2026-narrow-task-misalignment}'s
recipe and chat template. One-sided Fisher $p$, LoRA insecure against full-fine-tuning insecure.}
\label{tbl:coder}
\begin{tabular}{lccc}
\toprule
Cell & insecure & secure & $p$ \\
\midrule
LoRA (chat) & 16/400 (4.0\%) & 0/400 & --- \\
Full FT, chat & 1/160 & 0/160 & 0.024 \\
Full FT, chain-of-thought (plain) & 2/160 & 1/160 & 0.073 \\
Full FT, chain-of-thought (chat) & 2/160 & 0/160 & 0.073 \\
Full FT, plain & 0/160 & 2/160 & --- \\
\bottomrule
\end{tabular}
\end{table}

\subsection{Judge validation}\label{app:judge}

\begin{table}[H]
\revfloat
\centering\small
\caption{Judge validation. \textbf{Left:} our judge against GPT-4o on 678 responses from our GPT-4.1
replication, binarised at alignment $<30$ without the coherence gate; 22\% misaligned by GPT-4o, so
a judge that never flagged misalignment would agree on 77.7\%. Agreement 97.8\%, Cohen's
$\kappa=0.936$, PABAK 0.956, false-positive rate 1.5\% (8 of 527). \textbf{Right:} one of us rated
200 responses blind to the judge (two pages of 100, one from our Qwen2.5-32B cells and one from the
transplant, each stratified by condition and enriched for judge-flagged responses). Agreement 88\%
(176/200), $\kappa=0.741$, PABAK 0.76 ($\kappa$ 0.83 and 0.66 on the two pages).}
\label{tbl:judge}
\begin{tabular}{lcc@{\hspace{2.5em}}lcc}
\toprule
& \multicolumn{2}{c}{GPT-4o} & & \multicolumn{2}{c}{Human} \\
Our judge & misaligned & not & Our judge & misaligned & not \\
\midrule
misaligned & 144 & 8 & misaligned & 60 & 20 \\
not & 7 & 519 & not & 4 & 116 \\
\bottomrule
\end{tabular}
\end{table}

\subsection{LoRA learning-rate sweep}\label{app:lrsweep}

\begin{table}[H]
\revfloat
\centering\small
\caption{Ranks 1 and 64 at three learning rates (instruct, single seed). Broad EM is given over all
responses, with the within-coherent-prose rate in brackets; narrow is the insecure share of
coherent code. The decision rule, fixed before the runs, was that rank 1 exceeds rank 64 at every
learning rate (one-sided Fisher). It holds at $10^{-5}$ ($p=2\times10^{-16}$) and
$3\times10^{-5}$ ($p=9\times10^{-6}$) and fails at $3\times10^{-6}$, where rank 1 produces no
code and the reverse holds ($p=3\times10^{-13}$). The $10^{-5}$ rows are the rank ladder's runs.}
\label{tbl:lrsweep}
\begin{tabular}{lcccc}
\toprule
Cell & insecure & secure & narrow & final loss \\
\midrule
r1, $3\times10^{-6}$ & 0.00 (0.0) & 0.00 & no code & 0.254 \\
r1, $10^{-5}$ & 10.38 (15.8) & 0.25 & 10.3 & 0.198 \\
r1, $3\times10^{-5}$ & 3.75 (19.6) & 0.50 & 13.7 & 0.179 \\
r64, $3\times10^{-6}$ & 5.12 (8.1) & --- & 8.7 & 0.155 \\
r64, $10^{-5}$ & 1.25 (3.1) & --- & 10.2 & 0.145 \\
r64, $3\times10^{-5}$ & 0.62 (1.7) & --- & 15.3 & 0.142 \\
\bottomrule
\end{tabular}
\end{table}

\subsection{The refusal axis}\label{app:refusal}

\begin{table}[H]
\revfloat
\centering\small
\caption{Movement of the LoRA cells along the refusal direction of
\citet{arditi-et-al-2024-refusal-direction} and along $\hat P$ (raw residual units, mean
$\pm$ sd over seeds 0--4, EM probe, layers 49--53). $\cos(\text{refusal},\hat P)=+0.21$.}
\label{tbl:app-refusal}
\begin{tabular}{lcc}
\toprule
Cell & refusal axis & $\hat P$ \\
\midrule
instruct, insecure & $-2.4\pm0.9$ & $+7.9\pm1.2$ \\
instruct, secure & $-5.3\pm0.4$ & $+0.4\pm1.2$ \\
base, insecure & $+5.4\pm0.7$ & $+8.6\pm1.4$ \\
base, secure & $+2.2\pm0.4$ & $+0.2\pm0.8$ \\
\bottomrule
\end{tabular}
\end{table}

\subsection{An independently built persona vector}\label{app:vc}

The design and decision rules for this section were fixed before any result was seen, in a dated
design note in our repository (not a public registry). Table~\ref{tbl:vc-design} gives the
construction of $v_C$, Table~\ref{tbl:vc-causal} the causal tests, Table~\ref{tbl:g3} the drift
geometry on the independent probe, and Table~\ref{tbl:vc-outcomes} every prediction and its
outcome.

\begin{table}[H]
\revfloat
\centering\scriptsize\setlength{\tabcolsep}{3pt}
\caption{Construction of $v_C$, following \citet{chen-et-al-2025-persona-vectors} with their
released \texttt{evil} trait data. No fine-tuned model, no question of
\citeauthor{betley-et-al-2026-narrow-task-misalignment} and no EM judgement enters it.}
\label{tbl:vc-design}
\begin{tabular}{@{}p{0.24\linewidth}p{0.70\linewidth}@{}}
\toprule
Item & Setting \\
\midrule
Model & un-fine-tuned Qwen2.5-32B-Instruct, chat template with the system prompt \\
Prompts & five contrastive system-prompt pairs (evil vs.\ ethical) $\times$ twenty extraction questions; ten samples per polarity (1{,}000 each), temperature 1.0, 256 new tokens \\
Filter & positive rollouts with evil score $>50$, negative $<50$ (Chen et al.'s rule and judge prompt, on our judge); coherence $\ge50$; refusals excluded; 479 positive, 996 negative kept \\
Direction & difference of mean residual-stream activations over response tokens, per layer \\
Stability & question-grouped five-fold cross-validation AUC $\approx1.0$ (label-permuted 0.50); without the two extraction questions closest to Betley et al.'s, cosine 0.997 with $v_C$ \\
Injection scale & anchor $s^\star$: $v_C$'s injected norm equals $\hat P$'s in the transplant. Swept $\{0.25,0.5,1,2,4\}\times s^\star$ at $n{=}200$; chose the largest scale keeping coherent prose at $\ge70\%$ of the recipient's baseline share \\
Geometry & with $\hat P$: raw cosine 0.35--0.55 across $\hat P$ seeds 0--3; whitened 0.07--0.17, against $-0.03$ to $-0.13$ for a non-EM control \\
\bottomrule
\end{tabular}
\end{table}

\begin{table}[H]
\revfloat
\centering\small
\caption{$v_C$ injected into three recipients (layers 49--53, $n{=}800$ per condition; scale as a
multiple of $s^\star$). The random direction is norm-matched to $v_C$; one-sided Fisher $p$, $v_C$
vs.\ random. A norm-matched non-EM direction stays at the floor. The last row injects $\hat P$ into
the parent instruct model at $v_C$'s norm, in the same run.}
\label{tbl:vc-causal}
\begin{tabular}{lcccc}
\toprule
& \multicolumn{2}{c}{Betley et al.'s 8 questions} & \multicolumn{2}{c}{Held-out 20 questions} \\
Recipient (scale) & $v_C$ / random / baseline & $p$ & $v_C$ / random / baseline & $p$ \\
\midrule
raw base ($1\times s^\star$) & 3.88 / 0.88 / 0.25 & $5\times10^{-5}$ & 7.50 / 2.62 / 3.75 & $1\times10^{-5}$ \\
parent instruct ($2\times s^\star$) & 11.50 / 0.00 / 0.00 & $<10^{-5}$ & 2.88 / 0.00 / 0.00 & $<10^{-5}$ \\
secure-LoRA instruct ($1\times s^\star$) & 1.12 / 0.00 / 0.12 & 0.002 & 1.88 / 0.00 / 0.25 & $3\times10^{-5}$ \\
\midrule
parent instruct, $\hat P$ at $v_C$'s norm & 0.00 / 0.00 / 0.00 & --- & 0.12 / 0.00 / 0.00 & --- \\
\bottomrule
\end{tabular}
\end{table}

\begin{table}[H]
\revfloat
\centering\small
\caption{Drift cosines on the independent probe (s0, layers 49--53). Full-fine-tuning seeds 1--3
were not captured on this probe.}
\label{tbl:g3}
\begin{tabular}{lcc}
\toprule
Cell & $\cos(\Delta,\hat P)$ & $\cos(\Delta,v_C)$ \\
\midrule
base, LoRA r32 insecure & $+0.119$ & $+0.114$ \\
base, full FT insecure & $-0.177$ & $-0.121$ \\
instruct, LoRA r1 insecure & $-0.031$ & $-0.037$ \\
instruct, LoRA r8 insecure & $-0.031$ & $-0.038$ \\
instruct, LoRA r32 insecure & $-0.037$ & $-0.036$ \\
instruct, LoRA r64 insecure & $-0.079$ & $-0.074$ \\
instruct, full FT insecure & $-0.206$ & $-0.146$ \\
\bottomrule
\end{tabular}
\end{table}

\begin{table}[H]
\revfloat
\centering\scriptsize\setlength{\tabcolsep}{3pt}
\caption{Predictions fixed before the runs, and their outcomes.}
\label{tbl:vc-outcomes}
\begin{tabular}{@{}p{0.40\linewidth}p{0.54\linewidth}@{}}
\toprule
Prediction & Outcome \\
\midrule
$v_C$ overlaps $\hat P$ above a non-EM control & held, modestly (Table~\ref{tbl:vc-design}) \\
Drift on $v_C$: LoRA insecure $>0$, secure $\approx0$, full FT $<0$ & held in the base lineage; in the instruct lineage the contrasts held but the absolute signs failed (Table~\ref{tbl:g3}) \\
LoRA's weight update enriched on $v_C$, as on $\hat P$ & failed: every cell 1.04--1.49$\times$ \\
$v_C$ above random in every recipient and question set & held, 6 of 6 (Table~\ref{tbl:vc-causal}) \\
Ablating $v_C$ reduces the bad-medical broadcast & not supported: own direction 21.0\% $\to$ 10.1\% in the same run, $v_C$ 27.3\% (peak band, which is the last layer); all layers 7.6\% with coherent prose 26\% \\
LoRA rank 1 above rank 64 at every learning rate & failed at $3\times10^{-6}$ (Table~\ref{tbl:lrsweep}) \\
Bad-medical $\times$ LoRA above good-medical in both lineages & held (Table~\ref{tbl:inducer}) \\
Training-text directions above their own random directions & failed (Table~\ref{tbl:d4}) \\
\bottomrule
\end{tabular}
\end{table}

\subsection{Directions from the other inducers' training text}\label{app:d4}

\begin{table}[H]
\revfloat
\centering\small
\caption{Directions built in the un-fine-tuned instruct model from the inducers' training text alone
(bad minus good medical advice on the same 500 questions; risky financial or extreme-sports advice
minus the model's own answer to the same 500 prompts), injected into the raw base against their
own norm-matched random directions. Split-half stability 0.98--0.99; whitened cosine with $\hat P$
$-0.05$ to $0.14$. Baselines 0.25\% (Betley et al.'s questions) and 3.75\% (held-out). The primary
test, direction vs.\ its random direction at Bonferroni $\alpha=0.05/6$, fails for all three.}
\label{tbl:d4}
\begin{tabular}{lccc}
\toprule
Direction & Betley 8: direction / own random & $p$ & held-out 20: direction / own random \\
\midrule
medical & 2.00 / 1.00 & 0.074 & 3.75 / 3.62 \\
financial & 1.88 / 0.75 & 0.038 & 3.50 / 3.62 \\
sports & 2.12 / 0.75 & 0.017 & 3.88 / 4.00 \\
\bottomrule
\end{tabular}
\end{table}
\end{revblock}

\subsection{\rev{Part~IV mitigations at three seeds}}\label{app:mitig-seeds}

\begin{table}[H]
\revfloat
\centering\small
\caption{The Part~IV mitigations at three fine-tuning seeds (Qwen2.5-32B instruct, rs-LoRA rank~32
on insecure code; misaligned\% of 800, with the within-coherent-prose rate in brackets). At seeds 1
and 2, one-sided Fisher $p<10^{-5}$ for baseline $>$ inoculated, baseline $>$ persona-ablated and
random-ablated $>$ persona-ablated. Narrow: insecure share of coherent code on the code-security
evaluation (816 generations).}
\label{tbl:mitig-seeds}
\begin{tabular}{lccc}
\toprule
Cell & s0 & s1 & s2 \\
\midrule
rs-LoRA baseline & 4.75 (6.13) & 5.12 (6.83) & 5.50 (7.39) \\
Inoculated & 0.00 (0.00) & 0.00 (0.00) & 0.00 (0.00) \\
Persona-ablated & 1.62 (1.96) & 0.75 (0.82) & 1.12 (1.44) \\
Random-ablated & 4.38 (5.07) & 4.50 (5.49) & 6.00 (9.96) \\
\midrule
Narrow\%, baseline / inoculated & 27.8 / 9.7 & 26.3 / 11.0 & 26.7 / 9.7 \\
Code coherence\%, baseline / inoculated & 65.2 / 87.1 & 62.0 / 84.8 & 66.2 / 85.9 \\
\bottomrule
\end{tabular}
\end{table}

\subsection{\revb{Training-time steering at Qwen2.5-32B}}\label{app:d5}

\revb{The 7B training-time steering recipe of \S\ref{sec:medmech}, transposed to Qwen2.5-32B and fixed
before any run: bad-medical full fine-tuning of the instruct model at seeds 0--2 (the recipe of the
existing 32B bad-medical cell, which is the seed-0 unsteered run; seeds 1 and 2 have their own), with the 32B bad-medical persona direction subtracted from the
residual stream during training at layers 38--40 (the layers around the direction's held-out
within-AUC peak, the rule that gave layers 20--22 at 7B), at scale $\times\,\|d_l\|$ as at 7B; a
matched-norm random direction is the control. Steering is removed before evaluation.}

\begin{table}[H]
\revbfloat
\centering\small
\caption{The steering inversion at Qwen2.5-32B at three training seeds (misaligned\% of 800, with the
within-coherent-prose rate in brackets; each seed has its own unsteered run, the seed-0 one being the
existing 32B bad-medical full fine-tuning cell). One-sided Fisher on all $n$, steer-away at scale~4 $>$
own unsteered: $p=2.3\times10^{-5}$, $p<10^{-6}$ and $p=7.2\times10^{-5}$ at seeds 0, 1 and 2; random
$>$ own unsteered: $p\geq0.99$ at every seed; steer-away at scale~4 $>$ random: $p<10^{-6}$ at every
seed. Scale~2 was run at seed~0 only.}
\label{tbl:d5}
\begin{tabular}{lccc}
\toprule
Cell & s0 & s1 & s2 \\
\midrule
Unsteered & 22.1 (29.5) & 18.1 (23.7) & 17.3 (23.5) \\
Steer away, scale 2 & 33.5 (46.3) & --- & --- \\
Steer away, scale 4 & 31.3 (45.1) & 28.8 (40.7) & 25.1 (35.6) \\
Random direction, scale 4 & 17.8 (24.1) & 10.6 (12.5) & 7.8 (9.3) \\
\bottomrule
\end{tabular}
\end{table}

\revb{The second claim of \S\ref{sec:medmech}, that pushing full fine-tuning along the persona direction
does not install a broadcast, was repeated in the same way, also fixed before the run: insecure-code
full fine-tuning of the instruct model at seeds 0--2 (the recipe, generation settings and judge of the
existing 32B code cells, which are the unsteered runs), steered \emph{toward} the same 32B persona direction at layers 38--40, at scale~4
(and at scale~2 at seed~0), with a matched-norm random direction at scale~4, as in the 7B code arms.
Steering toward the persona is above its own unsteered run in no seed; pooled, it is $12/2400$ against
$12/2400$, a post hoc one-sided 95\% upper bound (Newcombe) of $0.35$ percentage points on any rise,
against $3.4\%$ for LoRA on the same data. Steering at
scale~4 makes the model answer in prose more often, along a random direction in every seed and along
the persona at seeds 0 and 1, but the misaligned share of that prose stays at the floor. Steering toward
the persona also reduces the narrow behaviour: pooled, the insecure share of coherent code is $15.9\%$
against $19.5\%$ unsteered (one-sided Fisher $p=0.04$), and $15.9\%$ is not clearly above the secure-code
control's $13.7\%$ ($p=0.12$). Most of the fall is at seed~2 ($15.3\%$ against $22.8\%$); at seeds 0 and 1 the
narrow rate is close to the unsteered run's, and there is still no broadcast. Because code full
fine-tuning produces no broadcast for steering to cancel, this tests installation, not the prevention
of \citet{chen-et-al-2025-persona-vectors}.}

\begin{table}[H]
\revbfloat
\centering\small
\caption{Code full fine-tuning steered toward the persona at Qwen2.5-32B at three training seeds
(misaligned\% of 800, with the within-coherent-prose rate in brackets; each seed's unsteered run is the
existing 32B code full fine-tuning cell for that seed). One-sided Fisher on all $n$, toward at scale~4
$>$ own unsteered: $p=0.11$, $0.86$ and $0.83$ at seeds 0, 1 and 2 (within coherent prose, $0.75$,
$>0.99$ and $0.76$); toward at scale~4 $>$ random: $p=0.73$, $0.50$ and $0.83$. At seed~0, toward at
scale~2 $>$ unsteered, $p=0.19$. Scale~2 was run at seed~0 only. Narrow: insecure share of coherent
code.}
\label{tbl:d6}
\begin{tabular}{lccc}
\toprule
Cell & s0 & s1 & s2 \\
\midrule
\multicolumn{4}{@{}l}{\emph{Misaligned\% (within coherent prose)}} \\
Unsteered & 0.12 (1.2) & 0.62 (6.3) & 0.75 (3.2) \\
Toward, scale 2 & 0.50 (3.6) & --- & --- \\
Toward, scale 4 & 0.62 (1.1) & 0.38 (1.1) & 0.50 (2.4) \\
Random direction, scale 4 & 0.75 (1.5) & 0.25 (0.6) & 0.75 (1.9) \\
\midrule
\multicolumn{4}{@{}l}{\emph{Coherent prose\% / narrow\%}} \\
Unsteered & 10.9 / 20.5 & 10.0 / 16.6 & 23.8 / 22.8 \\
Toward, scale 2 & 13.9 / 18.8 & --- & --- \\
Toward, scale 4 & 57.4 / 17.7 & 35.9 / 15.6 & 20.5 / 15.3 \\
Random direction, scale 4 & 50.4 / 21.0 & 43.6 / 14.0 & 40.4 / 20.9 \\
\bottomrule
\end{tabular}
\end{table}

\end{document}